\pdfoutput=1
\documentclass[11pt]{article}

\usepackage[]{emnlp2021}

\usepackage{times}
\usepackage{latexsym}
\usepackage{aligned-overset}
\usepackage{amsopn}
\usepackage{float}
\usepackage{enumitem}
\usepackage{booktabs}
\usepackage{tikz}
\usepackage{physics}
\usepackage{makecell}
\usepackage{bm}
\definecolor{aliceblue}{rgb}{0.94, 0.97, 1.0}
\definecolor{ugreen}{rgb}{0,0.5,0}
\usepackage{multirow}
\usepackage{threeparttable}
\usepackage{booktabs}
\usepackage{amsmath}

\usepackage[ruled,linesnumbered]{algorithm2e}

\usepackage[T1]{fontenc}
\usepackage[utf8]{inputenc}

\usepackage{microtype}

\title{RankNAS: Efficient Neural Architecture Search by Pairwise Ranking}

\author{
	Chi Hu$^1$,
	Chenglong Wang$^1$,
	Xiangnan Ma$^1$,
	Xia Meng$^1$, \\
	\textbf{Yinqiao Li$^1$,}
	\textbf{Tong Xiao$^{1,2}$\thanks{\xspace\xspace Corresponding author.},}
	\textbf{Jingbo Zhu$^{1,2}$}
	\textbf{Changliang Li$^{3}$} \\
	$^{1}$NLP Lab, School of Computer Science and Engineering \\ Northeastern University, Shenyang, China \\
	$^{2}$NiuTrans Research, Shenyang, China \\
	$^{3}$Kingsoft AI Lab, Beijing, China \\
	{\tt
		huchinlp@gmail.com, clwang1119@gmail.com,
		}\\
	{\tt
		\{xiaotong,zhujingbo\}@mail.neu.edu.cn
		} \\
}

\begin{document}
\maketitle
\begin{abstract}
This paper addresses the efficiency challenge of Neural Architecture Search (NAS) by formulating the task as a ranking problem. Previous methods require numerous training examples to estimate the accurate performance of architectures, although the actual goal is to find the distinction between “good” and “bad” candidates. Here we do not resort to performance predictors. Instead, we propose a performance ranking method (RankNAS) via pairwise ranking. It enables efficient architecture search using much fewer training examples. Moreover, we develop an architecture selection method to prune the search space and concentrate on more promising candidates. Extensive experiments on machine translation and language modeling tasks show that RankNAS can design high-performance architectures while being orders of magnitude faster than state-of-the-art NAS systems.
\end{abstract}

\section{Introduction}

Neural Architecture Search (NAS) has advanced state-of-the-art on various tasks, such as image classification \citep{dblp:conf/cvpr/zophvsl18,Pham2018EfficientNA,2018regularized, dblp:conf/cvpr/tancpvshl19}, machine translation \citep{DBLP:journals/taslp/FanTXQLL20, DBLP:conf/icml/SoLL19}, and language modeling \citep{ Pham2018EfficientNA, DBLP:conf/iclr/LiuSY19, jiang-etal-2019-improved, li-etal-2020-learning}. Despite the remarkable results, conventional NAS methods are computationally expensive, requiring training millions of architectures during search. For instance, obtaining a state-of-the-art machine translation model with an evolutionary algorithm requires more than 250 GPU years \citep{DBLP:conf/icml/SoLL19}. 

Several techniques have been proposed to improve the search efficiency, such as sharing parameters among all architectures \citep{Pham2018EfficientNA, DBLP:conf/aaai/CaiCZYW18, DBLP:conf/cvpr/ZhongYWSL18}, predicting the performance instead of full training \citep{ DBLP:conf/eccv/LiuZNSHLFYHM18, DBLP:conf/iclr/BakerGRN18, DBLP:conf/eccv/WenLCLBK20, DBLP:journals/corr/abs-2003-12857}, and searching over a continuous space \citep{ DBLP:conf/iclr/LiuSY19, jiang-etal-2019-improved, li-etal-2020-learning}. Unfortunately, these approaches still suffer from the high cost of predicting the performance of each candidate architecture. An inherent reason for this is that obtaining accurate performance requires training numerous neural networks to convergence, as described in Sec. \ref{sec:PE}. However, it is unnecessary to predict the model performance as in previous NAS methods. Rather, all we need is to distinguish architectures of different quality in NAS, say, ranking these architectures.

\begin{figure}[t!]
    \centering
    \includegraphics[scale=0.47]{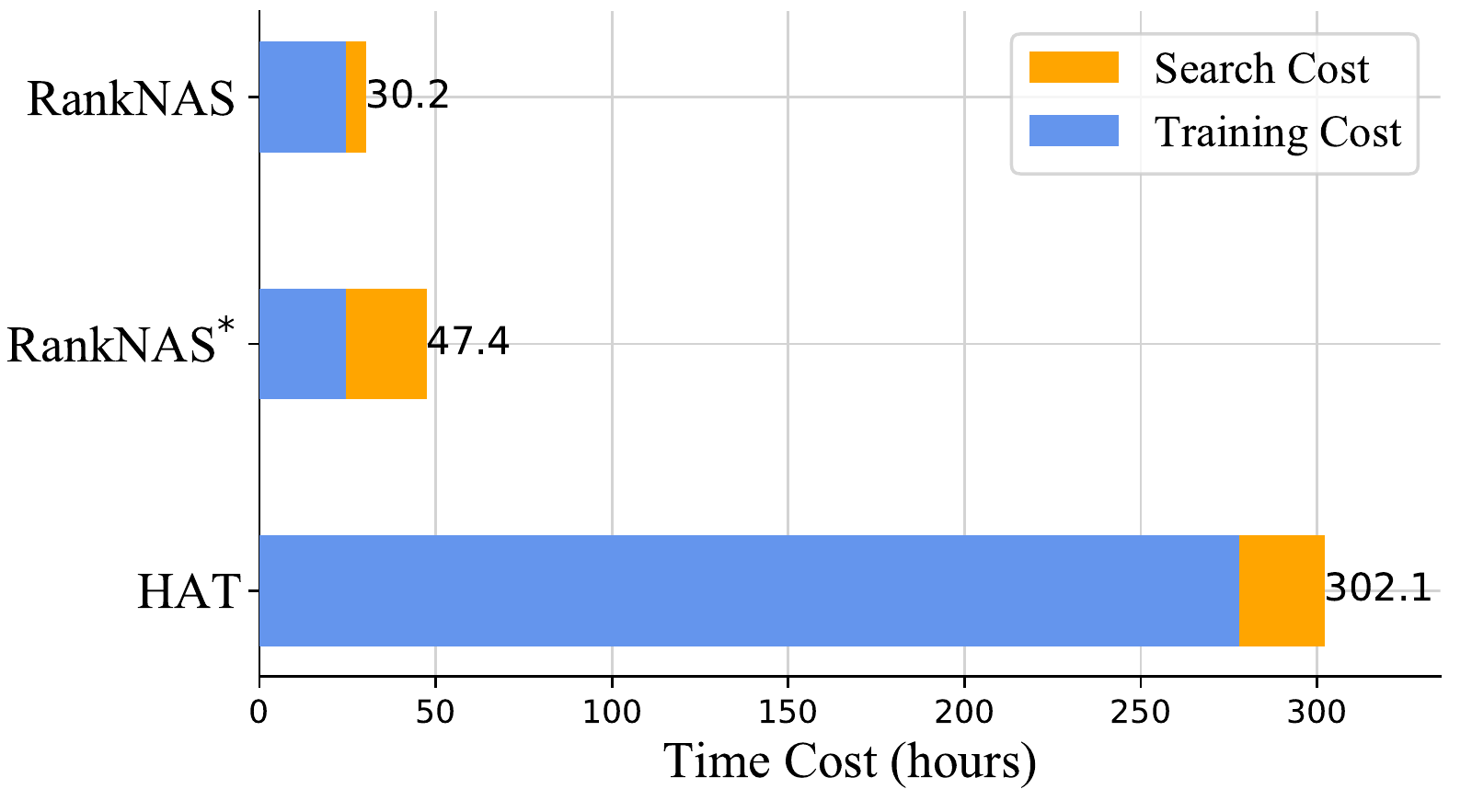}
    \caption{The time cost of different NAS methods on the WMT machine translation task. RankNAS{*} denotes the results without search space pruning. Our method significantly accelerates NAS through pairwise ranking and search space pruning.} 
    \label{fig:search-cost}
\end{figure}

In this paper, we approach the problem by formulating NAS as a ranking task. Here we propose RankNAS, a ranking model for comparing different architectures. One of the key challenges is that directly ranking all architectures in a large search space is still computationally infeasible. Therefore, we adopt the pairwise method \citep{DBLP:conf/icml/BurgesSRLDHH05, DBLP:conf/icml/WauthierJJ13}, where the ranking problem is reduced to a binary classification problem over architecture pairs. To speed up RankNAS further, we develop an architecture selection method that chooses the most promising architectures for evaluation according to the importance of features, e.g., the topology of architectures.

We test RankNAS on well-established machine translation and language modeling benchmarks. Experiments show that RankNAS is orders of magnitude faster than standard NAS systems and can find better architectures. Notably, RankNAS is generic to different tasks and evaluation metrics. It achieves competitive results on hardware-aware NAS tasks and is 10$\times$ faster than the HAT baseline \citep{wang2020hatht}. It also discovers new architectures that outperform vanilla Transformer by +1.8 BLEU points on the IWSLT'14 De-En data and +1.5 BLEU points on the WMT'14 En-De data, surpassing the Evolved Transformer \citep{DBLP:conf/icml/SoLL19} with 150,000$\times$ less search cost.

\section{Preliminaries}
NAS generally consists of two steps: 1) sample architectures from the pre-defined search space, and 2) estimate the performance of these samples. This work focuses on the performance estimation step, which is the efficiency bottleneck of NAS.

\subsection{Search Space}
\label{sec:search-space}
The search space contains all possible architectures for the search. In this work, we take the Transformer architecture for description, but the discussed problem and solutions are general and can be applied to other models. Following HAT \citep{wang2020hatht}, we represent a Transformer architecture as a set of features and search for the optimal model configuration. 

\begin{figure}[!t]
\centering
\begin{tikzpicture}
\definecolor{grannysmithapple}{rgb}{0.66, 0.89, 0.63}
\definecolor{flavescent}{rgb}{0.97, 0.91, 0.56}
\definecolor{babypink}{rgb}{0.96, 0.76, 0.76}
\definecolor{brightube}{rgb}{0.82, 0.62, 0.91}
\definecolor{columbiablue}{rgb}{0.61, 0.87, 1.0}
\definecolor{bubbles}{rgb}{0.91, 1.0, 1.0}
\definecolor{coral}{rgb}{1.0, 0.5, 0.31}
\definecolor{cornflowerblue}{rgb}{0.39, 0.58, 0.93}
\tikzstyle{fea} = [minimum width=1cm,minimum height=0.6cm,rectangle,inner sep=0.4em,rounded corners=3pt,draw,fill=gray!30]
\tikzstyle{head} = [minimum width=0.5cm,minimum height=0.3cm,rectangle,inner sep=0.2em,rounded corners=1pt,draw,fill=bubbles]
\tikzstyle{layer} = [minimum width=2.6cm,minimum height=0.7cm,rectangle,inner sep=0.2em,rounded corners=3pt,draw]
\tikzstyle{enbiglayer} = [minimum width=3cm,minimum height=5.2cm,rectangle,inner sep=0.4em,draw=gray!70,fill=grannysmithapple!60]
\tikzstyle{debiglayer} = [minimum width=3cm,minimum height=7cm,rectangle,inner sep=0.4em,draw=gray!70,fill=flavescent!40]
\tikzstyle{select} = [minimum width=2.65cm,minimum height=0.7cm,rectangle,inner sep=0.2em,rounded corners=3pt,draw,fill=white]
\tikzstyle{pre} = [minimum width=2cm,minimum height=1cm,rectangle,inner sep=0.2em,rounded corners=3pt,draw,fill=orange!20,align=center]
\tikzstyle{token} = [minimum width=0.5em,minimum height=0.5em,rectangle,inner sep=0.1em,align=center]
\begin{scope}
\node [layer] (enemb) at (0,0){};
\node [minimum width=3.5cm,minimum height=9.5cm,rectangle,rounded corners=3pt,draw=gray!50,thick,dotted](encoder)at ([yshift=10em]enemb.north){};
\node [layer,minimum width=2.1cm,fill=columbiablue!60] (enemb1) at (-0.25,0){};
\node at (0,0){\scriptsize{Embedding Dimension}};
\node [enbiglayer,anchor=north] (enlayer1) at ([yshift=14.2em]enemb.north) {};
\node [select,minimum width=1.5cm,minimum height=0.5cm,inner sep=0.1em,dashed] (norm11) at ([yshift=1.8em]enemb.north){\scriptsize{Norm}};
\node [select,minimum height=1cm] (enseat) at ([yshift=2.2em]norm11.north){};
\node at ([yshift=3.0em]norm11.north) {\scriptsize{Self-Attn Heads}};
%\draw[-]([yshift=0.3em]enseat.east) to ([yshift=0.3em]enseat.west);

\node [token](target) at ([xshift=-1.7em,yshift=-1.3em]enseat.north){\tiny{$t$}};
\node [token,fill=orange!20](t2) at ([xshift=-2.2em,yshift=-2.1em]enseat.north){};
\node [token,fill=orange!20](t3) at ([xshift=-1.7 em,yshift=-2.1em]enseat.north){};
\node [token,fill=orange!20](t4) at ([xshift=-1.2em,yshift=-2.1em]enseat.north){};
\node [token,fill=orange!10](t5) at ([xshift=-0.7em,yshift=-2.1em]enseat.north){};
\draw[-]([xshift=-2.2em,yshift=-2.1em]enseat.north) to ([xshift=-0.7em,yshift=-2.1em]enseat.north);
\draw[-]([xshift=-2.2em,yshift=-2.1em]enseat.north) to ([xshift=-2.2em,yshift=-1.95em]enseat.north);
\draw[-]([xshift=-1.7em,yshift=-2.1em]enseat.north) to ([xshift=-1.7em,yshift=-1.95em]enseat.north);
\draw[-]([xshift=-1.2em,yshift=-2.1em]enseat.north) to ([xshift=-1.2em,yshift=-1.95em]enseat.north);
\draw[-]([xshift=-0.7em,yshift=-2.1em]enseat.north) to ([xshift=-0.7em,yshift=-1.95em]enseat.north);
%\node [token,fill=orange!20](t2) at ([xshift=-2.2em,yshift=-2.1em]enseat.north){};
%\node [token,fill=orange!20](t3) at ([xshift=-1.7 em,yshift=-2.1em]enseat.north){};
%\node [token,fill=orange!20](t4) at ([xshift=-1.2em,yshift=-2.1em]enseat.north){};
%\node [token,fill=orange!10](t5) at ([xshift=-0.7em,yshift=-2.1em]enseat.north){};
\draw[-](target.south) to ([xshift=-2.2em,yshift=-1.85em]enseat.north);
\draw[-](target.south) to ([xshift=-1.7em,yshift=-1.85em]enseat.north);
\draw[-](target.south) to ([xshift=-1.2em,yshift=-1.85em]enseat.north);
\draw[-,dotted](target.south) to ([xshift=-0.7em,yshift=-1.85 em]enseat.north);
\node [head] at ([xshift=2.25em,yshift=-1.7em]enseat.north){\tiny{Head 2}};
\node [head] at ([xshift=1.75em,yshift=-1.95em]enseat.north){\tiny{Head 1}};
\node [select,minimum width=1.5cm,minimum height=0.5cm,inner sep=0.1em,fill=gray!20] (norm12) at ([yshift=1.5em]enseat.north){\scriptsize{Norm}};
\node [select] (ffn) at ([yshift=1.8em]norm12.north){};
\node [select,minimum width=1.6cm,fill=coral!50] (ffn1) at ([xshift=-1.4em,yshift=1.8em]norm12.north){};
\node at ([yshift=1.8em]norm12.north){\scriptsize{Feed-Forward Dimension}};
\node [select,minimum width=1.5cm,minimum height=0.5cm,inner sep=0.1em,dashed] (norm13) at ([yshift=1.5em]ffn.north){\scriptsize{Norm}};
\node at ([xshift=-2.4em,yshift=0.5em]norm13.north){\footnotesize{Layer 1}};

\node [layer,fill=grannysmithapple!60](enlayer2)at ([yshift=2.8em]norm13.north){\footnotesize{Layer 2}};
\node (endot)at ([yshift=1.2em]enlayer2.north){\footnotesize{$\textbf{\dots}$}};
\node at ([xshift=-1.2em,yshift=0.3em]endot.west){\scriptsize{\textbf {Fixed}}};
\node at ([xshift=-1.2em,yshift=-0.3em]endot.west){\scriptsize{Layer Num}};
\node [layer,fill=grannysmithapple!60](enlayerm)at ([yshift=1.6em]endot.north){\footnotesize{Layer $m$}};
\node at ([xshift=-3em,yshift=0.7em]enlayerm.north){\small{Encoder}};
\draw[->,thick](enemb.north) to (norm11.south);
\draw[->,thick](norm11.north) to (enseat.south);
\draw[->,thick](enseat.north) to (norm12.south);
\draw[->,thick](norm12.north) to (ffn.south);
\draw[->,thick](ffn.north) to (norm13.south);
\draw[->,thick](norm13.north) to (enlayer2.south);
\draw[->,thick](enlayer2.north) to (endot.south);
\draw[->,thick](endot.north) to (enlayerm.south);
\end{scope}
\begin{scope}[xshift=1.6in]
\node [layer] (deemb) at (0,0){};
\node [minimum width=3.5cm,minimum height=10.25cm,rectangle,rounded corners=3pt,draw=gray!50,thick,dotted](decoder)at ([yshift=11em]deemb.north){};
\node [layer,minimum width=2.1cm,fill=cornflowerblue!40] (deemb1) at (-0.25,0){};
\node at (0,0){\scriptsize{Embedding Dimension}};
\node [debiglayer,anchor=north] (delayer1) at ([yshift=18.85em]deemb.north) {};
\node [select,minimum width=1.5cm,minimum height=0.5cm,inner sep=0.1em,dashed] (norm21) at ([yshift=1.8em]deemb.north){\scriptsize{Norm}};
\node [select,minimum height=1cm] (deseat) at ([yshift=2.2em]norm21.north){};
\node at ([yshift=3em]norm21.north) {\scriptsize{Self-Attn Heads}};
\node [token](target2) at ([xshift=-1.7em,yshift=-1.3em]deseat.north){\tiny{$t$}};
\node [token,fill=columbiablue!60](t2) at ([xshift=-2.2em,yshift=-2.1em]deseat.north){};
\node [token,fill=columbiablue!60](t3) at ([xshift=-1.7 em,yshift=-2.1em]deseat.north){};
\node [token,fill=columbiablue!60](t4) at ([xshift=-1.2em,yshift=-2.1em]deseat.north){};
\node [token,fill=columbiablue!30](t5) at ([xshift=-0.7em,yshift=-2.1em]deseat.north){};
\draw[-]([xshift=-2.2em,yshift=-2.1em]deseat.north) to ([xshift=-0.7em,yshift=-2.1em]deseat.north);
\draw[-]([xshift=-2.2em,yshift=-2.1em]deseat.north) to ([xshift=-2.2em,yshift=-1.95em]deseat.north);
\draw[-]([xshift=-1.7em,yshift=-2.1em]deseat.north) to ([xshift=-1.7em,yshift=-1.95em]deseat.north);
\draw[-]([xshift=-1.2em,yshift=-2.1em]deseat.north) to ([xshift=-1.2em,yshift=-1.95em]deseat.north);
\draw[-]([xshift=-0.7em,yshift=-2.1em]deseat.north) to ([xshift=-0.7em,yshift=-1.95em]deseat.north);
\draw[-](target2.south) to ([xshift=-2.2em,yshift=-1.85em]deseat.north);
\draw[-](target2.south) to ([xshift=-1.7em,yshift=-1.85em]deseat.north);
\draw[-](target2.south) to ([xshift=-1.2em,yshift=-1.85em]deseat.north);
\draw[-,dotted](target2.south) to ([xshift=-0.7em,yshift=-1.85 em]deseat.north);
\draw[-,dotted](target.south) to ([xshift=-0.7em,yshift=-1.85 em]enseat.north);
\node [head,fill=brightube!40] at ([xshift=2.3em,yshift=-1.7em]deseat.north){\tiny{Head 2}};
\node [head,fill=brightube!40] at ([xshift=1.8em,yshift=-1.95em]deseat.north){\tiny{Head 1}};
\node [select,minimum width=1.5cm,minimum height=0.5cm,inner sep=0.1em,fill=gray!20] (norm22) at ([yshift=1.5em]deseat.north){\scriptsize{Norm}};
\node [select] (ende) at ([yshift=1.7em]norm22.north){};
\node at ([xshift=-1em,yshift=1.7em]norm22.north){\scriptsize{En-De Attn Heads}};
\node [head,fill=orange!20,inner sep=0.1em] at ([xshift=2.4em,yshift=-0.88em]ende.north){\tiny{Head 1}};
\node [select,minimum width=1.5cm,minimum height=0.5cm,inner sep=0.1em,fill=gray!20] (norm23) at ([yshift=1.5em]ende.north){\scriptsize{Norm}};
\node [select,fill=white] (ffn) at ([yshift=1.7em]norm23.north){};
\node [select,minimum width=1.2cm,fill=babypink!70] (ffn1) at ([xshift=-1.9em,yshift=1.7em]norm23.north){};
\node [fill=white] at ([xshift=-8.75em,yshift=-5.4em]ffn.east){\tiny{\textbf {Elastic}}};
\node [fill=white,inner sep=0.05em] at ([xshift=-8.75em,yshift=-6em]ffn.east){\tiny{En-De Attn}};
\node at ([yshift=1.7em]norm23.north){\scriptsize{Feed-Forward Dimension}};
\node [select,minimum width=1.5cm,minimum height=0.5cm,inner sep=0.1em,dashed] (norm24) at ([yshift=1.5em]ffn.north){\scriptsize{Norm}};
\node at ([xshift=-2.4em,yshift=0.5em]norm24.north){\footnotesize{Layer 1}};
\node (dedot)at ([yshift=2.2em]norm24.north){\footnotesize{$\textbf{\dots}$}};
\node at ([xshift=-1.2em,yshift=0.3em]dedot.west){\scriptsize{\textbf {Elastic}}};
\node at ([xshift=-1.2em,yshift=-0.3em]dedot.west){\scriptsize{Layer Num}};
\node [layer,fill=flavescent!40,draw,dotted,thick](delayern)at ([yshift=1.6em]dedot.north){\footnotesize{Layer $n$}};
\node at ([xshift=-3em,yshift=0.7em]delayern.north){\small{Decoder}};
\draw[->,thick](deemb.north) to (norm21.south);
\draw[->,thick](norm21.north) to (deseat.south);
\draw[->,thick](deseat.north) to (norm22.south);
\draw[->,thick](norm22.north) to (ende.south);
\draw[->,thick](ende.north) to (norm23.south);
\draw[->,thick](norm23.north) to (ffn.south);
\draw[->,thick](ffn.north) to (norm24.south);
\draw[->,thick](norm24.north) to (dedot.south);
%\draw[->,thick](delayer2.north) to (dedot.south);
\draw[->,thick](dedot.north) to (delayern.south);
\draw[->,line width=0.6pt](enlayerm.east) -- ([xshift=2.1em]enlayerm.east) -- ([xshift=-1.6em]ende.west) -- (ende.west);
\draw[-,line width=0.6pt,dashed](enlayer2.east) -- ([xshift=1.6em]enlayer2.east) -- ([xshift=-2.1em]ende.west) -- ([xshift=-1.6em]ende.west);
%\draw[-,line width=0.6pt]([xshift=-2.1em]ende.west) -- ([xshift=-1.6em]ende.west);
\end{scope}
\end{tikzpicture}

\caption{The architecture search space. We search for the optimal model size, e.g., the number of layers, and network topology, e.g., connections between different layers. The encoder part is ignored in the language modeling task. Appendix \ref{appendix-space} gives more details about the design choices for different tasks.} 
\label{fig:search-space} 
\end{figure}
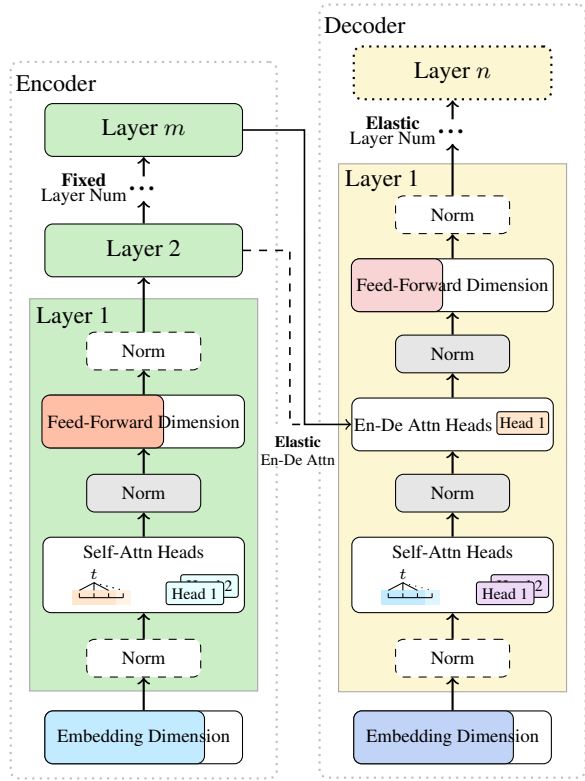

An overview of the search space is shown in Figure \ref{fig:search-space}. It is extended from the HAT's space and inspired by manually designed Transformer variants, including Relative Position Representations \cite{DBLP:conf/naacl/ShawUV18} and Deep Transformer \cite{wang-etal-2019-learning-deep}.  The search space can also be represented as a \textit{supernet} where each sub-network is a unique architecture. The search space contains around $10^{23}$ possible architectures, as detailed in Appendix \ref{appendix-space}. It is computationally prohibited to explore such a large space with an exhaustive method.

%----------------------------------------------
%----------------------------------------------

\subsection{Performance Estimation}
\label{sec:PE}
Let $\mathcal{A}$ denotes the search space, and each architecture in it is represented by a feature vector $\alpha$. Formally, the goal of NAS is to find the optimal architecture $\alpha^{*}$ with the best performance. The performance can be measured by some metrics, such as accuracy or latency. The performance estimation process consists of two steps: 1) estimate the performance of all architectures, and 2) choose the architecture with the optimal performance. 

Without loss of generality, we define $\mathcal{S}(\cdot)$ as the performance evaluated by some metrics. The task here is to find the most promising architecture with maximum $\mathcal{S}(\cdot)$. Standard NAS methods solve this problem by learning to estimate the performance of each architecture. The objective is given by:
\begin{equation}
\label{eq:nas}
\begin{aligned}
\alpha^{*}= \ & \operatorname*{argmax}_{\alpha} \mathcal{S}_{val}(w^{*}, \alpha)\\
\textrm{s.t.} \ w^{*}= \ & \operatorname*{argmax}_{w} \mathcal{S}_{train}(w, \alpha)
\end{aligned}
\end{equation}
where $w$ is the weights associated with the architecture. $\mathcal{S}_{val}$ and $\mathcal{S}_{train}$ are the evaluation results on the validation set and training set, respectively. 

Optimizing Eq. \ref{eq:nas} is time-consuming as obtaining the optimal weights for each architecture requires training them to converge. Although we can share the weights among all architectures to amortize the cost, performance evaluation is still nontrivial and requires numerous training steps.

\section{NAS as Ranking}
As mentioned in Sec. \ref{sec:PE}, the goal of NAS is to find promising architectures that achieve high performance on unseen data. NAS requires distinguishing whether the architectures are “good” or “bad” rather than predicting accurate performance. Therefore, it is natural to treat NAS as a ranking problem, in which the explicit goal is to rank different architectures correctly.

\subsection{Pairwise Ranking}
\paragraph{Problem Formulation.}
Given an architecture $\alpha$, we define a score $s$ on it by a function $r(\cdot)$:

\begin{equation}
s = r(\alpha, p)
\end{equation}

\noindent where $p$ is the parameter of the scoring function. We implement the scoring function with a gradient boosting decision tree, as detailed in Sec. \ref{sec:ranking-setup}.

We want to optimize $p$ such that $s$ assigns high scores to good architectures and low scores to bad architectures. This induces a ranking of the candidate architectures in the search space. It is infeasible to sort all candidate architectures in a large search space directly. A solution is to reduce the listwise ranking problem to the pairwise ranking problem. Fortunately, the properties of the NAS task allow us to achieve the goal. As described in \citet{dudziak2021brpnas}, the relation between any pair of performance is \textit{antisymmetric}, \textit{transitive} and \textit{connex}. This makes it possible to rank all architectures via pairwise comparisons, substantially reducing the training complexity.

\paragraph{Training Set Construction.}
\label{sec:traing-set}
In pairwise ranking, the learning task is framed as a \textit{binary classification} of architecture pairs into two categories: correctly ordered and incorrectly ordered. Given an architecture pair $(\alpha_{i}, \alpha_{j})$ and the order of performance $\bar{P}_{ij}$, we can construct training examples $(\alpha_{i}, \alpha_{j}, \bar{P}_{ij})$ for the classification by comparing the two values. Note that $\bar{P}_{ij}$ is a 0-1 variable. For example, if $\alpha_{i}$ is better than $\alpha_{j}$, we would add $(\alpha_{i}, \alpha_{j}, 1)$ and $(\alpha_{j}, \alpha_{i}, 0)$ to the training set.

\paragraph{Optimization.}
\label{sec:ranknas}
Consider a pair of architectures $(\alpha_{i}, \alpha_{j})$, scored by $s_i$ and $s_j$, respectively. The probability of $\alpha_{i}$ being better than $\alpha_{j}$ is given by the difference through an activation function $g$:
\begin{equation}
P_{i j}= {g\left(s_{i}-s_{j}\right)}
\end{equation}
We assume that $P_{i j}\geq0.5$ means $\alpha_{i}$ is better than $\alpha_{j}$ while $P_{i j}<0.5$ means $\alpha_{j}$ is better than $\alpha_{i}$. Here we use a logistic function to achieve this goal:
\begin{equation}
\label{eq:pij}
P_{i j}=\frac{1}{1+e^{-(s_{i}-s_{j})}}
\end{equation}
Similarly, $P_{j i}$ can be induced by:
\begin{equation}
\label{eq:pji}
P_{j i}= \frac{1}{1+e^{-(s_{j}-s_{i})}} = 1 - P_{i j}
\end{equation}
Denote the gold score of $\alpha_{i}$ being better than $\alpha_{j}$ as $\bar{P}_{i j}$. We use the cross-entropy loss function for the classification. The loss for a pair of inputs is:
\begin{equation}
\label{eq:ranknas}
\begin{aligned}
L_{i j} &=-(\bar{P}_{ij} \log P_{ij} + \bar{P}_{ji} \log P_{ji}) \\
&={(1-\bar{P}_{i j}) \cdot (s_{i}-s_{j})} + \\ 
& \ \ \ \ \ \log {(1+e^{-(s_{i}-s_{j})})}
\end{aligned}
\end{equation}

Compared with Eq. \ref{eq:nas}, Eq. \ref{eq:ranknas} just requires $\bar{P}_{i j}$. In particular, we use the intermediate performance measured on the validation set during training. It is much easier than assessing the accurate performance of candidate architectures. In this sense, the ranking model is ``easier'' to learn and may not need many training samples as in performance prediction. RankNAS also enables efficient optimization through gradient methods. Algorithm \ref{algo:training} describes the complete training process of the ranking model.

\begin{algorithm}[t]
\caption{Training of RankNAS}
\label{algo:training}
\KwIn {search space $\mathcal{A}$ and ranking model $r$}
\While{$r$ not converged}{
\textbf{training example construction:} sample $(\alpha_{i}, \alpha_{j})$ from $\mathcal{A}$, compute $\bar{P}_{ij}$ by comparing their performance\;
\textbf{classification:} compute scores $(s_i, s_j)$\;
\textbf{optimization:} optimize $r$ w.r.t. Eq. \ref{eq:ranknas}.
}
\end{algorithm}

\subsection{Applying Pairwise Ranking}
Although the training time of the ranking model is heavily reduced, it is still challenging to apply it to the ranking of all architectures in the search space $\mathcal{A}$. The challenge is that exploring all architectures is computationally expensive, even when the task is a binary classification.

%----------------------------------------------
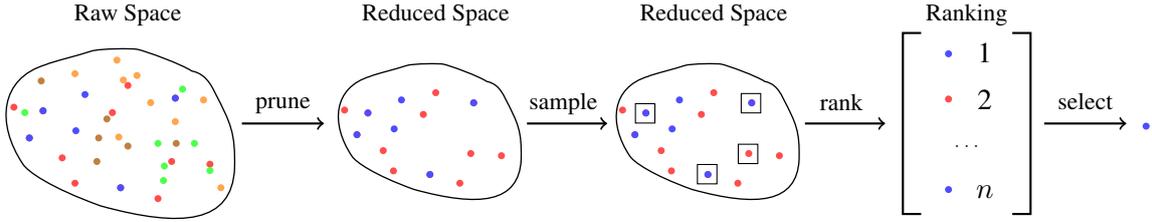
\begin{figure*}[t]
\definecolor{ublue}{rgb}{0.152,0.250,0.545}
\centering
\begin{tikzpicture}[scale=1.2]
\tikzstyle{sample} = [minimum width=0.05em,minimum height=0.05em,rectangle,draw]
\begin{scope}[xshift=-2in]
\draw [-,line width=0.5pt] (0,0.65)..controls (0.06,0.68) and (0.35,0.75)..(0.5,0.8)..controls (0.6,0.83) and (0.95,0.83)..(1,0.82)..controls (1.1,0.8) and (1.5,0.65)..(1.6,0.58)..controls (1.7,0.56) and (1.78,0.5)..(1.85,0.35)..controls (2.03,0.0) and (2.08,-0.1)..(2.07,-0.5)..controls (2.04,-1.1) and (1.5,-1.16)..(0.6,-0.91)..controls (0.05,-0.71) and (-0.2,-0.53)..(-0.25,-0.45)..controls (-0.55,0.0) and (-0.5,0.501)..(0,0.65);
\draw [->,thick](2.15,0)to(3.05,0);
\node at(2.6,0.2){\small{prune}};
\node at(0.9,1.2){\small{Raw Space}};
%%%%%%%%%RED
\node [anchor=center,red!70] at (0.18,-0.4){\Huge{$\cdot$}};
\node [anchor=center,red!70] at (0.9,0.4){\Huge{$\cdot$}};
\node [anchor=center,red!70] at (0.73,0.1){\Huge{$\cdot$}};
\node [anchor=center,red!70] at (-0.35,0.16){\Huge{$\cdot$}};
\node [anchor=center,red!70]at (-0.03,0.12){\Huge{$\cdot$}};
\node [anchor=center,red!70]at (-0.18,-0.18){\Huge{$\cdot$}};
\node [anchor=center,red!70]at (0.32,-0.68){\Huge{$\cdot$}};
\node [anchor=center,red!70] at (0.82,-0.73){\Huge{$\cdot$}};
\node [anchor=center,red!70]at (1.23,-0.85){\Huge{$\cdot$}};
\node [anchor=center,red!70] at (1.8,-0.47){\Huge{$\cdot$}};
\node [anchor=center,red!70]at (1.38,-0.44){\Huge{$\cdot$}};
\node [anchor=center,red!70]at (1.42,0.26){\Huge{$\cdot$}};
%%%%%%%%%%%%%
%%%%%%%%%BLUE
\node [anchor=center,blue!70] at (0.33,-0.1){\Huge{$\cdot$}};
\node [anchor=center,blue!70] at (0.43,0.3){\Huge{$\cdot$}};
\node [anchor=center,blue!70] at (-0.03,0.12){\Huge{$\cdot$}};
\node [anchor=center,blue!70] at (-0.18,-0.18){\Huge{$\cdot$}};
\node [anchor=center,blue!70] at (0.82,-0.73){\Huge{$\cdot$}};
\node [anchor=center,blue!70] at (1.42,0.26){\Huge{$\cdot$}};
%%%%%%%%%%%%%
%%%%%%%%%grEEN
%\node [anchor=center,green!70] at (1,0.3){\Huge{$\cdot$}};
\node [anchor=center,green!70]at (1.63,-0.24){\Huge{$\cdot$}};
\node [anchor=center,green!70]at (1.23,-0.24){\Huge{$\cdot$}};
\node [anchor=center,green!70]at (-0.23,0.1){\Huge{$\cdot$}};
\node [anchor=center,green!70]at (1.29,-0.65){\Huge{$\cdot$}};
\node [anchor=center,green!70] at (1.8,-0.54){\Huge{$\cdot$}};
\node [anchor=center,green!70]at (1.3,-0.49){\Huge{$\cdot$}};
\node [anchor=center,green!70]at (1.5,0.36){\Huge{$\cdot$}};
%%%%%%%%%%%%%
%%%%%%%%%orange
\node [anchor=center,orange!70] at (1.42,0){\Huge{$\cdot$}};
\node [anchor=center,orange!70] at (1.73,0.24){\Huge{$\cdot$}};
\node [anchor=center,orange!70] at (1.15,0.21){\Huge{$\cdot$}};
\node [anchor=center,orange!70]at (0.85,0.46){\Huge{$\cdot$}};
\node [anchor=center,orange!70]at (1,0.51){\Huge{$\cdot$}};
\node [anchor=center,orange!70]at (0.32,0.53){\Huge{$\cdot$}};
\node [anchor=center,orange!70] at (1.92,-0.73){\Huge{$\cdot$}};
\node [anchor=center,orange!70]at (0.78,0.68){\Huge{$\cdot$}};
\node [anchor=center,orange!70] at (0.8,-0.17){\Huge{$\cdot$}};
%%%%%%%%%%%%%
%%%%%%%%%brown
\node [anchor=center,brown]at (-0.05,0.45){\Huge{$\cdot$}};
\node [anchor=center,brown]at (1.43,-0.24){\Huge{$\cdot$}};
\node [anchor=center,brown] at (0.67,0.03){\Huge{$\cdot$}};
\node [anchor=center,brown]at (0.58,-0.18){\Huge{$\cdot$}};
\node [anchor=center,brown] at (0.9,-0.27){\Huge{$\cdot$}};
\node [anchor=center,brown]at (0.56,-0.44){\Huge{$\cdot$}};
%%%%%%%%%%%%%
\end{scope}

\begin{scope}[scale=0.8,xshift=-0.75in]
\draw [-,line width=0.5pt] (0,0.65)..controls (0.06,0.68) and (0.35,0.75)..(0.5,0.8)..controls (0.6,0.83) and (0.95,0.83)..(1,0.82)..controls (1.1,0.8) and (1.5,0.65)..(1.6,0.58)..controls (1.7,0.56) and (1.78,0.5)..(1.85,0.35)..controls (2.03,0.0) and (2.08,-0.1)..(2.07,-0.5)..controls (2.04,-1.1) and (1.5,-1.16)..(0.6,-0.91)..controls (0.05,-0.71) and (-0.2,-0.53)..(-0.25,-0.45)..controls (-0.55,0.0) and (-0.5,0.501)..(0,0.65);
\draw [->,thick](2.15,0)to(3.25,0);
\node at(2.65,0.28){\small{sample}};
\node at(0.9,1.5){\small{Reduced Space}};
%%%%%%%%%RED
\node [anchor=center,red!70] at (0.18,-0.4){\Huge{$\cdot$}};
\node [anchor=center,red!70] at (0.9,0.4){\Huge{$\cdot$}};
\node [anchor=center,red!70] at (0.73,0.1){\Huge{$\cdot$}};
\node [anchor=center,red!70] at (-0.35,0.16){\Huge{$\cdot$}};
\node [anchor=center,red!70]at (-0.03,0.12){\Huge{$\cdot$}};
\node [anchor=center,red!70]at (-0.18,-0.18){\Huge{$\cdot$}};
\node [anchor=center,red!70]at (0.32,-0.68){\Huge{$\cdot$}};
\node [anchor=center,red!70] at (0.82,-0.73){\Huge{$\cdot$}};
\node [anchor=center,red!70]at (1.23,-0.85){\Huge{$\cdot$}};
\node [anchor=center,red!70] at (1.8,-0.47){\Huge{$\cdot$}};
\node [anchor=center,red!70]at (1.38,-0.44){\Huge{$\cdot$}};
\node [anchor=center,red!70]at (1.42,0.26){\Huge{$\cdot$}};
%%%%%%%%%%%%%
%%%%%%%%%BLUE
\node [anchor=center,blue!70] at (0.33,-0.1){\Huge{$\cdot$}};
\node [anchor=center,blue!70] at (0.43,0.3){\Huge{$\cdot$}};
\node [anchor=center,blue!70] at (-0.03,0.12){\Huge{$\cdot$}};
\node [anchor=center,blue!70] at (-0.18,-0.18){\Huge{$\cdot$}};
\node [anchor=center,blue!70] at (0.82,-0.73){\Huge{$\cdot$}};
\node [anchor=center,blue!70] at (1.42,0.26){\Huge{$\cdot$}};
%%%%%%%%%%%%%
\end{scope}

\begin{scope}[scale=0.8,xshift=0.75in]
\draw [-,line width=0.5pt] (0,0.65)..controls (0.06,0.68) and (0.35,0.75)..(0.5,0.8)..controls (0.6,0.83) and (0.95,0.83)..(1,0.82)..controls (1.1,0.8) and (1.5,0.65)..(1.6,0.58)..controls (1.7,0.56) and (1.78,0.5)..(1.85,0.35)..controls (2.03,0.0) and (2.08,-0.1)..(2.07,-0.5)..controls (2.04,-1.1) and (1.5,-1.16)..(0.6,-0.91)..controls (0.05,-0.71) and (-0.2,-0.53)..(-0.25,-0.45)..controls (-0.55,0.0) and (-0.5,0.501)..(0,0.65);
\draw [->,thick](2.15,0)to(3.25,0);
\node at(2.65,0.29){\small{rank}};
\node at(0.9,1.5){\small{Reduced Space}};
\node [sample] at(1.37,-0.42) {};
\node [sample] at(-0.04,0.14) {};
\node [sample] at(0.81,-0.7) {};
\node [sample] at(1.41,0.28) {};
%%%%%%%%%RED
\node [anchor=center,red!70] at (0.18,-0.4){\Huge{$\cdot$}};
\node [anchor=center,red!70] at (0.9,0.4){\Huge{$\cdot$}};
\node [anchor=center,red!70] at (0.73,0.1){\Huge{$\cdot$}};
\node [anchor=center,red!70] at (-0.35,0.16){\Huge{$\cdot$}};
\node [anchor=center,red!70]at (-0.03,0.12){\Huge{$\cdot$}};
\node [anchor=center,red!70]at (-0.18,-0.18){\Huge{$\cdot$}};
\node [anchor=center,red!70]at (0.32,-0.68){\Huge{$\cdot$}};
\node [anchor=center,red!70] at (0.82,-0.73){\Huge{$\cdot$}};
\node [anchor=center,red!70]at (1.23,-0.85){\Huge{$\cdot$}};
\node [anchor=center,red!70] at (1.8,-0.47){\Huge{$\cdot$}};
\node [anchor=center,red!70]at (1.38,-0.44){\Huge{$\cdot$}};
\node [anchor=center,red!70]at (1.42,0.26){\Huge{$\cdot$}};
%%%%%%%%%%%%%
%%%%%%%%%BLUE
\node [anchor=center,blue!70] at (0.43,0.3){\Huge{$\cdot$}};
\node [anchor=center,blue!70] at (0.33,-0.1){\Huge{$\cdot$}};
\node [anchor=center,blue!70] at (-0.03,0.12){\Huge{$\cdot$}};
\node [anchor=center,blue!70] at (-0.18,-0.18){\Huge{$\cdot$}};
\node [anchor=center,blue!70] at (0.82,-0.73){\Huge{$\cdot$}};
\node [anchor=center,blue!70] at (1.42,0.26){\Huge{$\cdot$}};
%%%%%%%%%%%%%
\end{scope}

\begin{scope}[xshift=1.7in]
\node at(0.7,1.2){\small{Ranking}};
\draw [->,thick](1.55,0)to(2.45,0);
\node at(2.0,0.25){\small{select}};
\node [anchor=center,blue!70] at (0.5,0.75){\Huge{$\cdot$}};
\node at (0.9,0.78){\normalsize{1}};
\node [anchor=center,red!70] at (0.5,0.25){\Huge{$\cdot$}};
\node at (0.9,0.27){\normalsize{2}};
\node [anchor=center] at (0.7,-0.25){\tiny{$\mathbf \dots$}};
%\node at (0.9,-0.25){\scriptsize{1}};
\node [anchor=center,blue!70] at (0.5,-0.75){\Huge{$\cdot$}};
\node at (0.9,-0.75){\normalsize{$n$}};
\draw [-,thick] (0.2,1) -- (0,1) -- (0,-1) -- (0.2,-1);
\draw [-,thick] (1.2,1) -- (1.4,1) -- (1.4,-1) -- (1.2,-1);
\end{scope}

\begin{scope}[xshift=2.75in]
\node [anchor=center,blue!70] at (0,-0.05){\Huge{$\cdot$}};
\end{scope}
\end{tikzpicture}
\caption{The proposed search process consists of three steps: 1) prune the search space according to the importance of architectural features, 2) sample $n$ architectures from the \textit{reduced search space} by specific strategies, and 3) rank them with the trained ranking model and choose the best one. Here different color means different features.} 
\label{fig:overview}
\end{figure*}
%----------------------------------------------
\begin{figure}[ht]
    \centering
    \includegraphics[scale=0.47]{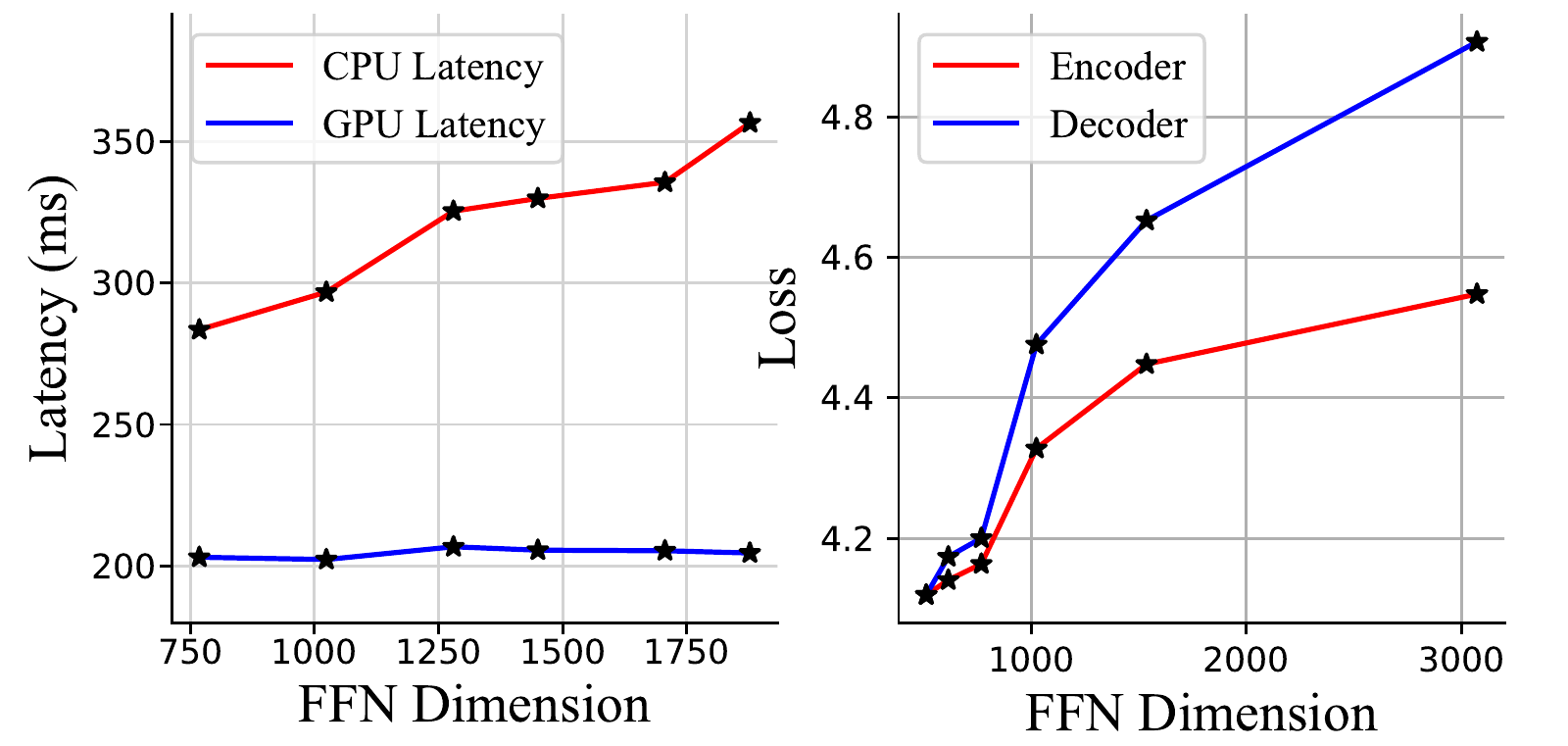}
    \caption{The impact of FFN dimension on latency and validation loss. All results are obtained on the WMT’14 En-De task with the same settings described in Sec.  \ref{exp:hat}.}
    \label{fig:performance-features}
\end{figure}

\paragraph{Correlations between Features and Performance.}
We start by analyzing the effect of architectural features on estimated performance. Figure \ref{fig:performance-features} illustrates the impact of the FFN dimension on latency and the validation loss on the machine translation task. We observe that: (a) different architectural features have very different correlations with the same evaluation metric, and (b) the same features also have different influences on different metrics. For example, the latency monotonically increases when scaling the FFN dimension on CPUs, while it is almost unchanged on GPUs. Hence, it is natural to improve search efficiency by eliminating unimportant features.

\paragraph{Feature Importance.}
\label{sec:feature-selection}
Inspired by previous feature selection methods \citep{DBLP:journals/ml/Breiman01, DBLP:journals/jmlr/FisherRD19}, we measure the importance of an architectural feature (e.g., the number of layers) by calculating the increase in the model error after permuting the feature.

We assume that each architecture $\alpha$ is represented by a feature vector $\bm{f} \in \mathcal{R}^{M \times N}$, where $M$ is the number of different features, and $N$ is the dimension of feature vectors. Also, we assume a set $C$ that contains $n$ architectures sampled from the search space. We first estimate the original model error $L_{total}$ on $C$ using the accumulation of the prediction errors. For any feature $\bm{f}_i \in \bm{f}$, we randomize it for each architecture in $C$. Then the randomized architectural features are passed to the ranking model and yield an error $L_i$. The importance of the $i$-th feature $\bm{f}_i$ is defined by:
\begin{equation}
    I(\bm{f}_i) = \frac{L_{i}}{L_{total}}
\end{equation}
where a higher value implies $\bm{f}_i$ is more important.

\paragraph{Search Space Pruning.}
It is easy to select valuable architectural features with the above measure. Given all features $\bm{f} \in \mathcal{R}^{M \times N}$, we discard those with a score less than a threshold $\theta$ and obtain the selected features $\bm{f'} \in \mathcal{R}^{M' \times N}$, where $M' < M$. Then we can prune the search space according to the selected features. For instance, if the feature \textit{Embedding Dimension} is not selected, we will keep it \textit{fixed} during the search. Finally, we only search over the architectures in the reduced search space.

An overview of the search process is presented in Figure \ref{fig:overview}. As described in Sec. \ref{sec:ranknas}, the training of the proposed ranking model is much cheaper than previous methods, which need to optimize the parameters for all architectures. Pruning search space further reduces the number of architectures to be evaluated. Also, the sampling procedure can be implemented with any existing NAS search strategy, e.g., Random Search (RS) or Evolution Algorithm (EA).

\section{Experiments}
\label{sec:experiment}
\subsection{Experimental Setups}
We evaluate our methods on language modeling and machine translation tasks. In the experiments, we search for hardware-aware architectures and high-accuracy architectures.

%----------------------------------------------------------------------------------------------------
\renewcommand{\arraystretch}{1.0}
\begin{table*}[!t]
\centering
\begin{tabular}{c r l r r r r r}
\toprule[1pt]
\rule{0pt}{10pt}
\multirow{2}{*}{Hardware}& 
\multirow{2}{*}{Task} & 
\multicolumn{1}{c}{\multirow{2}{*}{Method}} & 
\multicolumn{1}{c}{\multirow{2}{*}{\shortstack[c]{Latency \\ (ms)}}} & 
\multicolumn{1}{c}{\multirow{2}{*}{\#Params}}&
\multicolumn{1}{c}{\multirow{2}{*}{\shortstack[c]{FLOPs \\\\ (G)}}} & 
\multicolumn{1}{c}{\multirow{2}{*}{BLEU}} & 
\multicolumn{1}{c}{\multirow{2}{*}{\shortstack[c]{Search Cost\\\\ (hours)}}} \\ 
 & & & & & & & \\
\midrule 
\multirow{6}{*}{\shortstack[c]{Intel Xeon\\\\Silver 4114\\\\CPU}}&  
\multirow{3}{*}{WMT}&
Transformer&1031.4&213.0M&12.7&28.4&-\cr
 & &HAT&396.8&\textbf{67.9M}&4.2&28.5&335.1 \cr
 & &RankNAS&\textbf{384.2}&68.1M&\textbf{4.0}&\textbf{28.6}&\textbf{31.8} \cr
\cmidrule(lr){2-8}
&\multirow{4}{*}{IWSLT}&
Transformer&353.5&34.9M&1.6&34.4&-\cr
 & &HAT&\textbf{190.5}&\textbf{27.9M}&\textbf{1.4}&34.5&31.7 \cr
 & &RankNAS&197.4&29.6M&1.5&\textbf{34.6}&\textbf{7.2} \cr
\midrule 
\multirow{6}{*}{\shortstack[c]{NVIDIA\\\\GTX 1080Ti\\\\GPU}} & 
\multirow{3}{*}{WMT}&
Transformer&249.6&213.0M&12.7&28.4&-\cr
 & &HAT&214.8&66.2M&4.1&\textbf{28.5}&302.1 \cr
 & &RankNAS&\textbf{201.7}&\textbf{62.1M}&\textbf{3.9}&28.4&\textbf{30.2} \cr
\cmidrule(lr){2-8}
&\multirow{4}{*}{IWSLT}&
Transformer&200.9&34.9M&1.6&34.4&-\cr
 & &HAT&159.4&\textbf{33.9M}&1.6&34.7&24.5 \cr
 & &RankNAS&\textbf{148.2}&35.4M&\textbf{1.4}&\textbf{34.7}&\textbf{5.8} \cr
\bottomrule[1pt]
\end{tabular}
\caption{Comparisons of latency, model size, FLOPs, BLEU, and the overall search cost on machine translation tasks for the standard Transformer, HAT, and discovered architectures by our method. We mark the best results in bold for all metrics. Search costs are measured on a single RTX 2080Ti GPU.}
\label{tab:hat-search}
\end{table*}
%----------------------------------------------------------------------------------------------------

\paragraph{Training Setups.}
For machine translation, we experiment on the IWSLT’14 De-En and WMT’14 En-De tasks using the identical settings as \citet{wang2020hatht}. For language modeling, we experiment on the WikiText-103 dataset \citep{Merity2017PointerSM} with the same settings as \citet{baevski2019adaptiveir}. We set the maximum number of tokens per sample to 1,843 to fit the memory constraints and apply gradient accumulation to keep the same batch size as \citet{baevski2019adaptiveir} ’s work. All models are trained with mixed precision on 8 NVIDIA RTX 2080 Ti GPUs except for IWSLT ones, which only take one GPU for training. 

\paragraph{Ranking Model Setups.}
\label{sec:ranking-setup}
We implement the ranking model (binary classifier) described in Sec.\ref{sec:ranknas} with LightGBM \citep{DBLP:conf/nips/KeMFWCMYL17} and set the learning rate to 0.1. To prevent overfitting, we set the maximum number of leaves to 30 and the tree's maximum depth to 6. We also use the default regularization terms and apply the early stopping strategy to the training. Specifically, the training stops if the validation score does not increase for 5 rounds. After training the ranking model, we apply the search space pruning method to find the most valuable architectural features for different tasks and hardware. There are two hyper-parameters for pruning: the sample size and the threshold. We set them to 200/1.15 and 300/1.25 for the hardware-aware architecture search and high-accuracy architecture search, respectively. 

%----------------------------------------------------------------------------------------------------
\renewcommand{\arraystretch}{1.0}
\begin{table}[!t]
\centering
\setlength{\tabcolsep}{1mm}{ 
\begin{tabular}{l c c c }
\toprule[1pt]
\multicolumn{1}{c}{\multirow{2}{*}{Method}}& 
\multirow{2}{*}{\makecell[c]{Latency \\(CPU)}} & 
\multirow{2}{*}{\makecell[c]{Latency \\(GPU)}} & 
\multirow{2}{*}{PPL}\\
 & & & \\
\midrule
\citet{baevski2019adaptiveir}&\multicolumn{1}{r}{12.49}&0.53&18.70\\
\citet{dai2019transformerxlal}&\multicolumn{1}{r}{11.23}&0.42&18.30\\
\citet{DBLP:conf/acl/PressSL20}&\multicolumn{1}{r}{12.17}&0.52&\textbf{17.96}\\
\midrule
RankNAS\ (Ours)&\multicolumn{1}{r}{\textbf{4.83}}&\textbf{0.29}&18.13\\
\bottomrule[1pt]
\end{tabular}}
\caption{Performance of our discovered model and the state-of-the-art language models. The perplexities are evaluated on the WikiText-103 test data. Latency is measured in units of seconds. All models have a similar size, around 250M.}
\label{tab:lm-search}
\end{table}

\paragraph{Architecture Search Setups.}
Table \ref{tab:iwslt-search-space} and Table \ref{tab:wmt-search-space} presents the search space of high-accuracy search for the translation tasks. We refer the readers to \citet{wang2020hatht}'s work for more details about the search space of hardware-aware architecture search. RankNAS is not restricted to a specific search strategy. We compare different search strategies in the experiments, including Random Search (RS) and Evolutionary Algorithm (EA). We apply uniform sampling for RS and use the same settings as \citet{wang2020hatht}'s work for EA. More specifically, the random search process will stop if the best-so-far architecture does not change for 3 epochs.

\renewcommand{\arraystretch}{1.0}
\begin{table*}[!t]
\centering
\begin{tabular}{l c c c | c c c}
\toprule[1pt]
 &\multicolumn{3}{c}{IWSLT'14 De-En}&\multicolumn{3}{c}{WMT'14 En-De}
\\
\midrule[1pt]
\multicolumn{1}{c}{\multirow{2}{*}{Method}}& 
\multicolumn{1}{c}{\multirow{2}{*}{\#Params}} & 
\multicolumn{1}{c}{\multirow{2}{*}{BLEU}} & 
\multicolumn{1}{c |}{\multirow{2}{*}{\shortstack[c]{Search Cost \\ \\ (hours)}}} & 
\multicolumn{1}{c}{\multirow{2}{*}{\#Params}} & 
\multicolumn{1}{c}{\multirow{2}{*}{BLEU}} & 
\multicolumn{1}{c}{\multirow{2}{*}{\shortstack[c]{Search Cost \\ \\ (hours)}}} \\ 
 & & & & & & \\ 
\midrule[1pt]
\citet{DBLP:conf/nips/VaswaniSPUJGKP17} & 35M & 34.4 & -& 213M & 28.4 & -\\
\citet{DBLP:conf/naacl/ShawUV18} & 37M & 35.4 & -& 213M & 29.2 & -\\
\citet{DBLP:conf/iclr/WuFBDA19} & 43M & 35.2 & -& 213M & 29.7 & -\\
% \citet{DBLP:journals/corr/abs-2008-00623} & 30M & 35.3 & - & - & - & -\\
\midrule
\citet{DBLP:conf/aaai/PhamL21} & 37M & 35.8 & - & - & - & - \\
\citet{DBLP:conf/icml/SoLL19} & - & - & - & 218M & 29.8 & 5.5$\times10^{6}$ \\
\citet{DBLP:journals/taslp/FanTXQLL20} & 38M & 36.1 & 262.7 & 213M & \textbf{30.1} & 1970.3 \\
\citet{DBLP:conf/acl/ZhaoDSZWC21} & - & - & - & 215M & 29.8 & 798.3 \\
RankNAS\ (Ours) & \textbf{34M} & \textbf{36.2} & \textbf{2.3} & \textbf{202M} & 29.9 & \textbf{36.9}\\
\bottomrule
\end{tabular}
\caption{Results on the IWSLT'14 De-En and WMT'14 En-De machine translation tasks. The models above are both designed by human experts, while the models below are discovered by NAS. Search costs are normalized to GPU hours on a single RTX 2080Ti GPU, according to the results on public benchmarks\footnote{\url{https://lambdalabs.com/gpu-benchmarks}}.}
\label{tab:accuracy-search}
\end{table*}

\paragraph{Evaluation Metrics.}
We report the results obtained by averaging 5 runs with different seeds. We calculate BLEU scores with case-sensitive tokenization using Moses, and apply the compound splitting BLEU for WMT, the same as HAT. We test the latency of models on an Intel Xeon Silver 4114 CPU and an NVIDIA GTX 1080Ti GPU. A machine translation model's latency is the time of translating a single sentence with a fixed length - 30 for WMT and 23 for IWSLT. For language modeling, the latency is the cost of decoding a single sentence without mini-batching, averaged over the whole test set. Following \citet{wang2020hatht}'s work, we measure each model's latency 300 times and remove the fastest and slowest 10\% and then take the average of the rest 80\%. Note that we report the total number of trainable parameters in a model, while \citet{wang2020hatht} emit the parameters of the embedding layers. The search cost is the GPU hours measured on or normalized to a single RTX 2080Ti.

\subsection{Results}

\paragraph{Hardware-Aware Architecture Search.}
\label{exp:hat}
The hardware-aware NAS aims to maximize the accuracy under specified latency constraints on different hardware platforms. We first rank architectures by their latencies and pick those that meet the constraint to achieve this goal. Then we rank the selected architectures by their losses on the validation set and choose the best one. For machine translation tasks, we use the same search space as HAT \citep{wang2020hatht}, which contains around $10^{15}$ possible architectures. For the language modeling task, we use the following search space: [10, 12, 14] for decoder layer number, [768, 1024] for embedding dimension, [3072, 4096, 5120] for hidden dimension, and [8, 12, 16] for the head number in attention modules. We add a simple linear projection without bias if two adjacent layers have different hidden sizes.  

Table \ref{tab:hat-search} shows the results of RankNAS comparing to HAT \citep{wang2020hatht} and Transformer \citep{DBLP:conf/nips/VaswaniSPUJGKP17} on the machine translation tasks. Our method is effective in reducing the search cost for different tasks and hardware platforms. For instance, it requires 10.53$\times$ less cost to find a comparable architecture on the WMT task. The discovered architectures also have the lowest latencies with the same or better BLEU scores on most tasks. For example, the architecture designed for the CPU is 2.68$\times$ faster than the standard Transformer.

We present the architecture search results for language modeling on the WikiText-103 test data in Table \ref{tab:lm-search}. All models are evaluated with a context window of 2,560 tokens, following \citet{baevski2019adaptiveir}. Our method significantly accelerates the baseline on different devices. Specifically, our method speeds up the baseline by 2.59$\times$ on the CPU and 1.83$\times$ on the GPU. Our model also obtains a perplexity of 18.13, which outperforms Transformer-XL \citep{dai2019transformerxlal} and is comparable to the state-of-the-art language model, e.g., Sandwich-Transformer \citep{DBLP:conf/acl/PressSL20}.

\paragraph{High-Accuracy Architecture Search.}
\label{exp:accuracy-search}
Unlike hardware-aware architecture search, the high-accuracy architecture search only optimizes accuracy and does not consider latency. In the experiments, we enlarge the HAT's search space by introducing two additional features \textit{Relative Attention Position} \citep{DBLP:conf/naacl/ShawUV18} and \textit{Layer Norm Position}, as shown in Table \ref{tab:iwslt-search-space} and Table \ref{tab:wmt-search-space}. This expands the size of search space to $10^{23}$, 8 orders of magnitude larger than HAT. 

%----------------------------------------------------------------------------------------------------
\renewcommand{\arraystretch}{1.0}
\begin{table}[!t]
\centering
\setlength{\tabcolsep}{1.2mm}{
\begin{tabular}{c c c c }
\toprule[1pt]
\multirow{2}{*}{\shortstack[c]{Search \\\\ Space}}&
\multirow{2}{*}{Method}&
\multirow{2}{*}{{Kendall’s $\bm{\tau}$}}&
\multirow{2}{*}{{Spearman’s $\bm{\rho}$}}\\
 & & & \\
\midrule
\multirow{2}{*}{Small}&  
HAT&0.827&0.913 \cr 
&Ours&\textbf{0.883}&\textbf{0.945}\cr
\midrule 
\multirow{2}{*}{Large}&  
HAT&0.754&0.842 \cr 
&Ours&\textbf{0.826}&\textbf{0.907}\cr
\bottomrule[1pt]
\end{tabular}}
\caption{RankNAS vs. HAT in terms of Kendall and Spearman rank correlation coefficient. The results are collected using the settings described in Sec. \ref{analysis:accuracy}.}
\label{tab:mt-predictor}
\end{table}
%----------------------------------------------------------------------------------------------------

We compare RankNAS with state-of-the-art machine translation models designed by human experts and models discovered by other NAS methods. The results are presented in Table \ref{tab:accuracy-search}. RankNAS consistently outperforms other methods in both the IWSLT and WMT tasks. It demonstrates that RankNAS can also design high-accuracy architectures. Notably, the discovered architectures achieve a +1.8 BLEU improvement on the IWSLT task and a +1.5 BLEU improvement on the WMT task than the standard Transformers baseline \citep{DBLP:conf/nips/VaswaniSPUJGKP17}. We show that RankNAS surpasses the Evolved Transformer \citep{DBLP:conf/icml/SoLL19}, with orders of magnitude fewer search costs. RankNAS also matches the performance of gradient-based methods, including NAO \citep{DBLP:journals/taslp/FanTXQLL20} and DARTSformer \citep{DBLP:conf/acl/ZhaoDSZWC21}. 

\section{Analysis}
We analyze both the accuracy and efficiency of our search method and study the effect of different features on model performance.

\subsection{Architecture Ranking Accuracy}
\label{analysis:accuracy}
To study the accuracy of the proposed method, we evaluate it on the IWSLT translation task. We randomly sample 200 different architectures from the HAT search space (small) and the enlarged search space (large) introduced in Sec. \ref{exp:accuracy-search}. We train these architectures from scratch and measure their BLEU scores on the test set. Table \ref{tab:mt-predictor} presents the Kendall and Spearman rank correlation coefficient between the predicted results and the real scores. RankNAS outperforms HAT in terms of different ranking correlations. For example, RankNAS achieves a high Kendall's Tau of 0.883 and 0.826 on small and large spaces. This indicates that the predicted ranking is very close to the real results.

\begin{figure}[ht]
    \centering
    \includegraphics[scale=0.47]{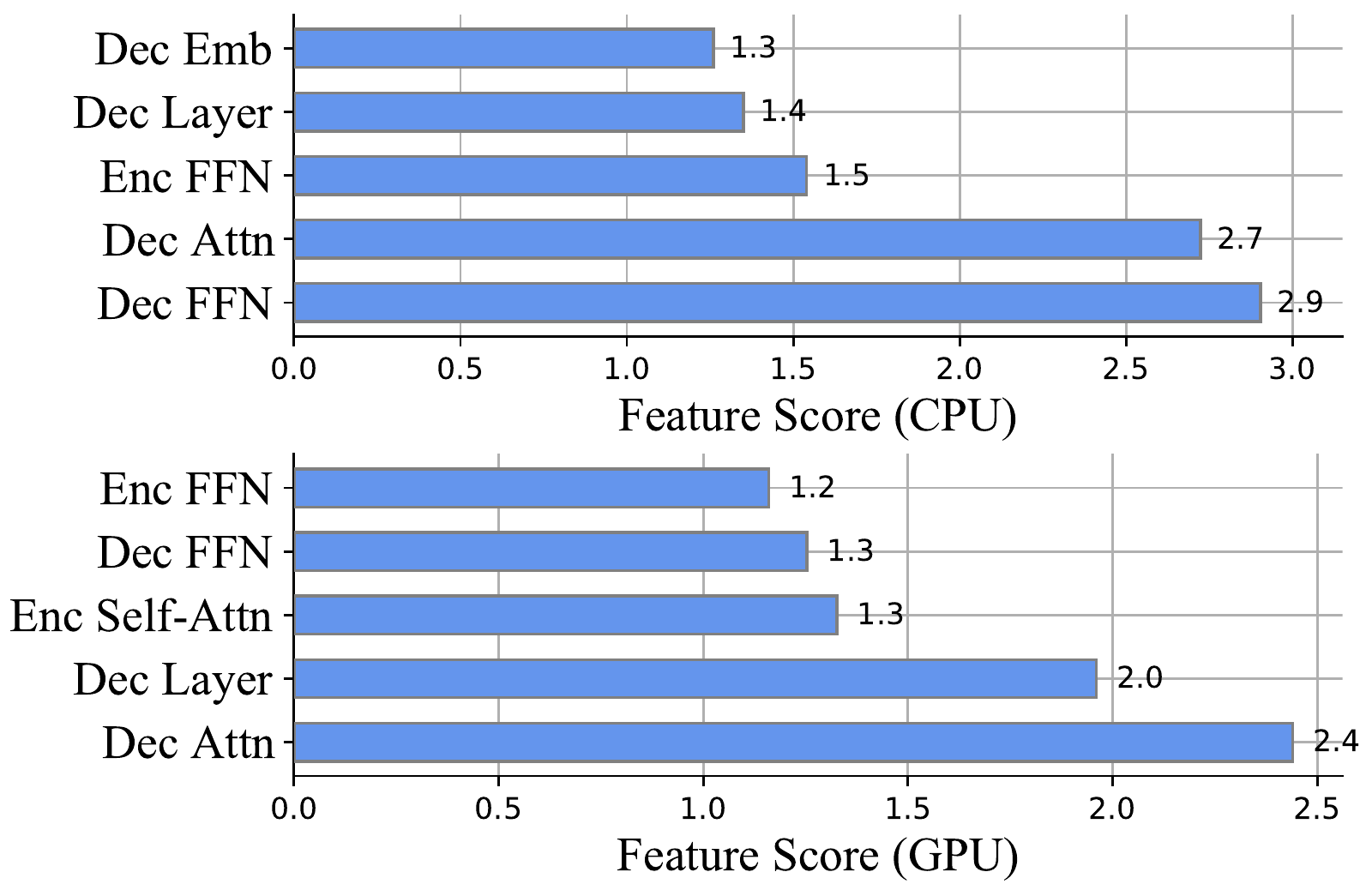}
    \caption{The selected features for different hardware platforms. A higher score means the feature is more important than others.}
    \label{fig:selected-features}
\end{figure}

\begin{figure}[t!]
    \centering
    \includegraphics[scale=0.47]{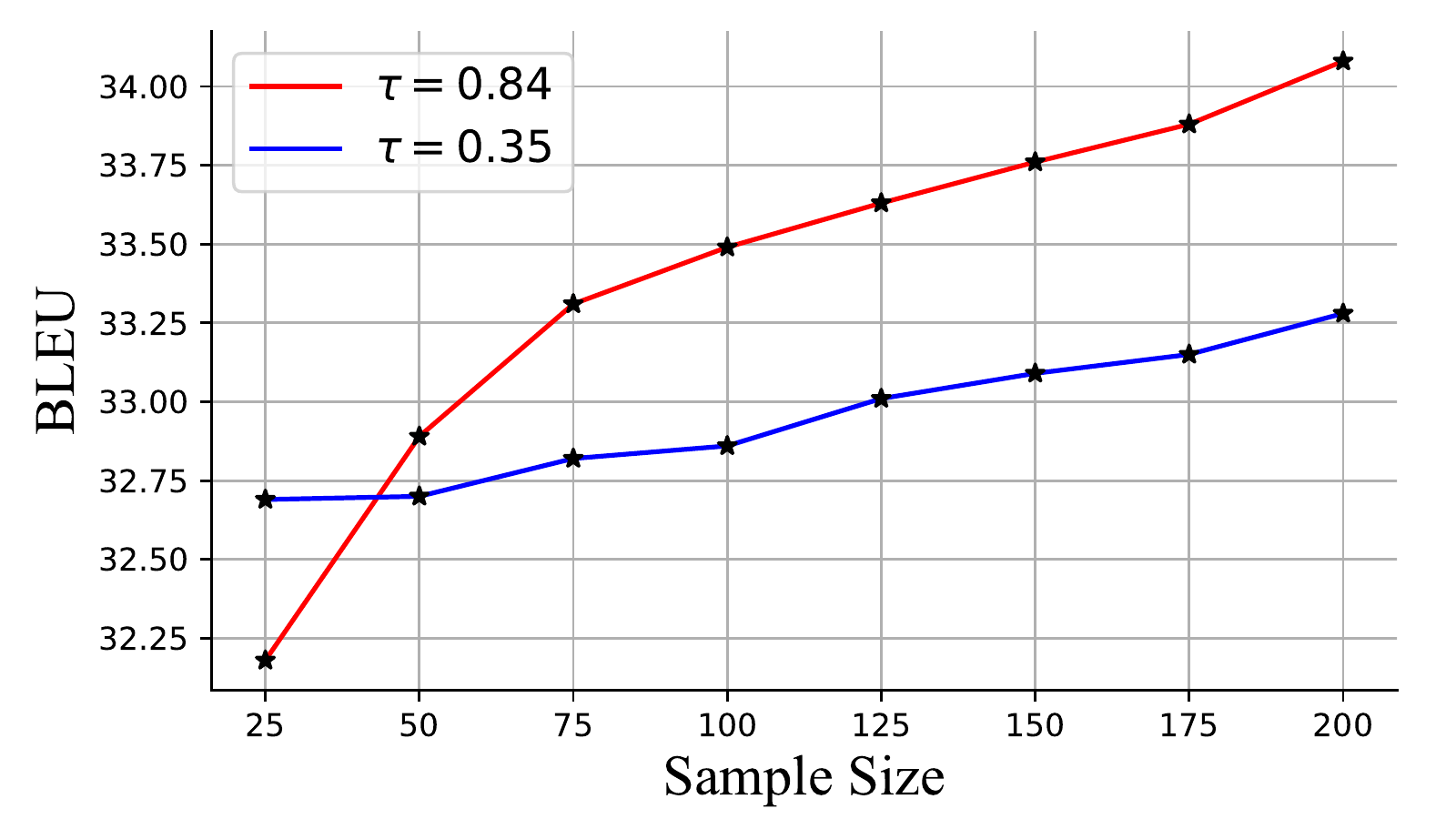}
    \caption{Search results of different ranking models. The inaccurate ranking model (in blue) leads to worse search results than the accurate ranking model (in red).}
    \label{fig:bleu-sample}
\end{figure}

\paragraph{Importance of Ranking Accuracy.}
\label{sec:analysis-accuracy}
Although our ranking model is more accurate than prior methods, a question remains: how does ranking accuracy affect the search quality? We analyze the impact of different ranking models on the high-accuracy NAS task. Figure \ref{fig:bleu-sample} compares two ranking models with different ranking correlation coefficients. The results are obtained by best-so-far models trained from scratch on the IWSLT'14 De-En data. Results show that inaccurate ranking leads to poor search results. It indicates that an accurate ranking model is essential for architecture search.

\subsection{Analysis of Discovered Architectures}
We present the discovered architectures in Appendix \ref{appendix-arch} and analyze important features for different hardware on the IWSLT'14 De-En task.

Figure \ref{fig:selected-features} (top) plots the selected features for the CPU. It shows that the decoder FFN dimension is the most important feature for predicting latency, followed by the decoder's arbitrary attention and the encoder FFN dimension. We also find that the decoder embedding dimension has a similar impact on latency as the number of decoder layers. 

Figure \ref{fig:selected-features} (bottom) illustrates the results for the GPU. Similar to the CPU, the latency on the GPU has a high correlation to the decoder attention module. The main difference is that the latency on GPU is insensitive to FFN or embedding dimensions but more sensitive to the number of decoder layers. 

The results indicate that we can design \textit{``shallow and wide''} models for GPUs and \textit{``deep and thin''} models for CPUs to achieve the Pareto-optimal state. Similar design insights have been verified in recent works, such as \citet{wang-etal-2019-learning-deep}, \citet{hu-etal-2020-niutrans}, \citet{DBLP:conf/aaai/LiLXZ21}, and \citet{lin-etal-2021-weight}.

\subsection{Search Efficiency}
Experiments in Sec. \ref{sec:experiment} show that our method has much lower search costs than previous works. We now analyze how does our method accelerates the architecture search.

\paragraph{Ranking Model Training Efficiency.}
The overall search cost includes the training time of the ranking model and the cost of the search process. Figure \ref{fig:search-cost} compares our method and HAT on the IWSLT’14 De-En task. The two methods share the same search space and sampling strategy for search. We observe that the ranking model training takes most of the time. RankNAS speeds up the ranking model training by 10.34 times compared with HAT. Pruning the search space further reduces the 75\% time of the search process. Thus the overall search cost is significantly reduced. It indicates that efficient training of the ranking model is essential to accelerate the search process.

\begin{figure}[t!]
    \centering
    \includegraphics[scale=0.47]{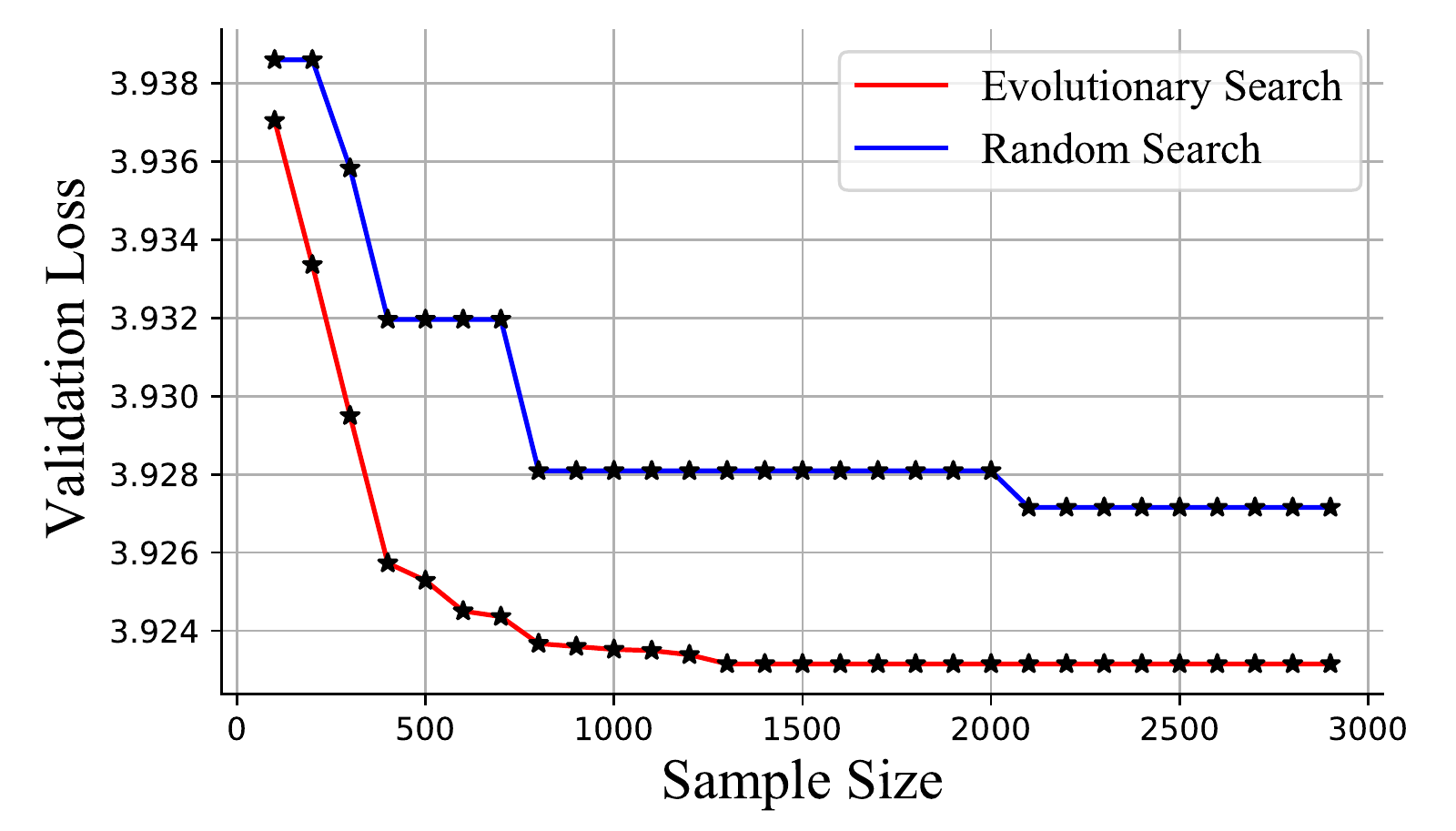}
    \caption{RankNAS combined with an evolutionary strategy achieves faster convergence and better results than other search methods.}
    \label{fig:loss}
\end{figure}

\paragraph{Architecture Search Efficiency.}
We also analyze the efficiency of our proposed methods on the IWSLT hardware-aware task. Figure \ref{fig:loss} shows the loss curves on the validation set of the models found by our method with different sampling strategies. We observe that RankNAS is compatible with different strategies. Also, the evolutionary algorithm outperforms random search in terms of the rate of convergence and the search result. 

\section{Related Work}
Many efforts have been made to improve the NAS efficiency for different tasks \citep{dblp:conf/cvpr/tancpvshl19, dblp:journals/corr/abs-1812-03443, dblp:journals/corr/abs-1812-00332, DBLP:journals/corr/abs-1911-00105, Chen2020AdaBERTTB}. A common approach to accelerating the search process is to use a proxy, such as reduced model size, training data, or training steps. However, it is inaccurate for estimating the model's performance and diminishes the NAS quality \citep{DBLP:conf/iclr/BakerGRN18,dudziak2021brpnas}. Another popular way is to share parameters among all architectures to reduce the training time \citep{dblp:conf/cvpr/tancpvshl19, dblp:journals/corr/abs-1812-00332}. However, it is infeasible to train all architecture candidates fairly to obtain their accurate performance. 

Recent works explored performance prediction based on architectural properties, i.e., the network topology and the model size \citep{DBLP:conf/eccv/LiuZNSHLFYHM18, 8901943, DBLP:conf/eccv/WenLCLBK20, DBLP:conf/eccv/NingZZWY20}. For instance, Hardware-Aware Transformer (HAT) \citep{wang2020hatht} encoded architectures into feature vectors and predicted the latency with a Multilayer Perceptron (MLP) for the target hardware. BRP-NAS \citep{dudziak2021brpnas} proposed an end-to-end performance predictor based on a Graph Convolutional Network (GCN). Although these methods greatly improve the performance estimation efficiency, they still require many samples and train numerous neural networks to converge, thereby increasing the search cost. Instead, we are motivated by the fact that NAS is expected to distinguish different candidate architectures. Thus, NAS can be solved by learning pairwise ranking rather than obtaining the accurate performance of architectures.

\section{Conclusion}
We have presented RankNAS, a simple yet efficient NAS algorithm for both hardware-aware and high-accuracy architecture search. We have shown that pairwise ranking can significantly improve search efficiency. We also have proposed a search space pruning method to help the ranking model be more efficient during the search. Our approach outperforms prior methods in both efficiency and accuracy. RankNAS requires 80\% less time in ranking model training on the hardware-aware search task and accelerates the overall search process by 11.53 times. Also, the architectures discovered by our method outperform state-of-the-art Transformer models in terms of efficiency and accuracy.

\section*{Acknowledgements}
This work was supported in part by the National Science Foundation of China (Nos. 61876035 and 61732005), the National Key R\&D Program of China (No.2019QY1801), and the Ministry of Science and Technology of the PRC (Nos. 2019YFF0303002 and 2020AAA0107900). The authors would like to thank the anonymous reviewers for their comments and suggestions.

\bibliography{anthology,custom}

\begin{thebibliography}{42}
\expandafter\ifx\csname natexlab\endcsname\relax\def\natexlab#1{#1}\fi

\bibitem[{Baevski and Auli(2019)}]{baevski2019adaptiveir}
Alexei Baevski and Michael Auli. 2019.
\newblock \href {https://openreview.net/forum?id=ByxZX20qFQ} {Adaptive input
  representations for neural language modeling}.
\newblock In \emph{7th International Conference on Learning Representations,
  {ICLR} 2019, New Orleans, LA, USA, May 6-9, 2019}. OpenReview.net.

\bibitem[{Baker et~al.(2018)Baker, Gupta, Raskar, and
  Naik}]{DBLP:conf/iclr/BakerGRN18}
Bowen Baker, Otkrist Gupta, Ramesh Raskar, and Nikhil Naik. 2018.
\newblock \href {https://openreview.net/forum?id=HJqk3N1vG} {Accelerating
  neural architecture search using performance prediction}.
\newblock In \emph{6th International Conference on Learning Representations,
  {ICLR} 2018, Vancouver, BC, Canada, April 30 - May 3, 2018, Workshop Track
  Proceedings}. OpenReview.net.

\bibitem[{Breiman(2001)}]{DBLP:journals/ml/Breiman01}
Leo Breiman. 2001.
\newblock \href {https://doi.org/10.1023/A:1010933404324} {Random forests}.
\newblock \emph{Mach. Learn.}, 45(1):5--32.

\bibitem[{Burges et~al.(2005)Burges, Shaked, Renshaw, Lazier, Deeds, Hamilton,
  and Hullender}]{DBLP:conf/icml/BurgesSRLDHH05}
Christopher J.~C. Burges, Tal Shaked, Erin Renshaw, Ari Lazier, Matt Deeds,
  Nicole Hamilton, and Gregory~N. Hullender. 2005.
\newblock \href {https://doi.org/10.1145/1102351.1102363} {Learning to rank
  using gradient descent}.
\newblock In \emph{Machine Learning, Proceedings of the Twenty-Second
  International Conference {(ICML} 2005), Bonn, Germany, August 7-11, 2005},
  volume 119 of \emph{{ACM} International Conference Proceeding Series}, pages
  89--96. {ACM}.

\bibitem[{Cai et~al.(2018)Cai, Chen, Zhang, Yu, and
  Wang}]{DBLP:conf/aaai/CaiCZYW18}
Han Cai, Tianyao Chen, Weinan Zhang, Yong Yu, and Jun Wang. 2018.
\newblock \href
  {https://www.aaai.org/ocs/index.php/AAAI/AAAI18/paper/view/16755} {Efficient
  architecture search by network transformation}.
\newblock In \emph{Proceedings of the Thirty-Second {AAAI} Conference on
  Artificial Intelligence, (AAAI-18), the 30th innovative Applications of
  Artificial Intelligence (IAAI-18), and the 8th {AAAI} Symposium on
  Educational Advances in Artificial Intelligence (EAAI-18), New Orleans,
  Louisiana, USA, February 2-7, 2018}, pages 2787--2794. {AAAI} Press.

\bibitem[{Cai et~al.(2019)Cai, Zhu, and
  Han}]{dblp:journals/corr/abs-1812-00332}
Han Cai, Ligeng Zhu, and Song Han. 2019.
\newblock \href {https://openreview.net/forum?id=HylVB3AqYm} {Proxylessnas:
  Direct neural architecture search on target task and hardware}.
\newblock In \emph{7th International Conference on Learning Representations,
  {ICLR} 2019, New Orleans, LA, USA, May 6-9, 2019}. OpenReview.net.

\bibitem[{Chen et~al.(2020)Chen, Li, Qiu, Wang, Li, Ding, Deng, Huang, Lin, and
  Zhou}]{Chen2020AdaBERTTB}
Daoyuan Chen, Yaliang Li, Minghui Qiu, Zhen Wang, Bofang Li, Bolin Ding, Hongbo
  Deng, Jun Huang, Wei Lin, and Jingren Zhou. 2020.
\newblock \href {https://doi.org/10.24963/ijcai.2020/341} {Adabert:
  Task-adaptive {BERT} compression with differentiable neural architecture
  search}.
\newblock In \emph{Proceedings of the Twenty-Ninth International Joint
  Conference on Artificial Intelligence, {IJCAI} 2020}, pages 2463--2469.
  ijcai.org.

\bibitem[{Dai et~al.(2019)Dai, Yang, Yang, Carbonell, Le, and
  Salakhutdinov}]{dai2019transformerxlal}
Zihang Dai, Zhilin Yang, Yiming Yang, Jaime Carbonell, Quoc Le, and Ruslan
  Salakhutdinov. 2019.
\newblock \href {https://doi.org/10.18653/v1/P19-1285} {Transformer-{XL}:
  Attentive language models beyond a fixed-length context}.
\newblock In \emph{Proceedings of the 57th Annual Meeting of the Association
  for Computational Linguistics}, pages 2978--2988, Florence, Italy.
  Association for Computational Linguistics.

\bibitem[{Dudziak et~al.(2020)Dudziak, Chau, Abdelfattah, Lee, Kim, and
  Lane}]{dudziak2021brpnas}
Lukasz Dudziak, Thomas C.~P. Chau, Mohamed~S. Abdelfattah, Royson Lee, Hyeji
  Kim, and Nicholas~D. Lane. 2020.
\newblock \href
  {https://proceedings.neurips.cc/paper/2020/hash/768e78024aa8fdb9b8fe87be86f64745-Abstract.html}
  {{BRP-NAS:} prediction-based {NAS} using gcns}.
\newblock In \emph{Advances in Neural Information Processing Systems 33: Annual
  Conference on Neural Information Processing Systems 2020, NeurIPS 2020,
  December 6-12, 2020, virtual}.

\bibitem[{Fan et~al.(2020)Fan, Tian, Xia, Qin, Li, and
  Liu}]{DBLP:journals/taslp/FanTXQLL20}
Yang Fan, Fei Tian, Yingce Xia, Tao Qin, Xiang{-}Yang Li, and Tie{-}Yan Liu.
  2020.
\newblock \href {https://doi.org/10.1109/TASLP.2020.2995270} {Searching better
  architectures for neural machine translation}.
\newblock \emph{{IEEE} {ACM} Trans. Audio Speech Lang. Process.},
  28:1574--1585.

\bibitem[{Fisher et~al.(2019)Fisher, Rudin, and
  Dominici}]{DBLP:journals/jmlr/FisherRD19}
Aaron Fisher, Cynthia Rudin, and Francesca Dominici. 2019.
\newblock \href {http://jmlr.org/papers/v20/18-760.html} {All models are wrong,
  but many are useful: Learning a variable's importance by studying an entire
  class of prediction models simultaneously}.
\newblock \emph{J. Mach. Learn. Res.}, 20:177:1--177:81.

\bibitem[{Hu et~al.(2020)Hu, Li, Li, Lin, Li, Wang, Xiao, and
  Zhu}]{hu-etal-2020-niutrans}
Chi Hu, Bei Li, Yinqiao Li, Ye~Lin, Yanyang Li, Chenglong Wang, Tong Xiao, and
  Jingbo Zhu. 2020.
\newblock \href {https://doi.org/10.18653/v1/2020.ngt-1.24} {The {N}iu{T}rans
  system for {WNGT} 2020 efficiency task}.
\newblock In \emph{Proceedings of the Fourth Workshop on Neural Generation and
  Translation}, pages 204--210, Online. Association for Computational
  Linguistics.

\bibitem[{Jiang et~al.(2019)Jiang, Hu, Xiao, Zhang, and
  Zhu}]{jiang-etal-2019-improved}
Yufan Jiang, Chi Hu, Tong Xiao, Chunliang Zhang, and Jingbo Zhu. 2019.
\newblock \href {https://doi.org/10.18653/v1/D19-1367} {Improved differentiable
  architecture search for language modeling and named entity recognition}.
\newblock In \emph{Proceedings of the 2019 Conference on Empirical Methods in
  Natural Language Processing and the 9th International Joint Conference on
  Natural Language Processing (EMNLP-IJCNLP)}, pages 3585--3590, Hong Kong,
  China. Association for Computational Linguistics.

\bibitem[{Ke et~al.(2017)Ke, Meng, Finley, Wang, Chen, Ma, Ye, and
  Liu}]{DBLP:conf/nips/KeMFWCMYL17}
Guolin Ke, Qi~Meng, Thomas Finley, Taifeng Wang, Wei Chen, Weidong Ma, Qiwei
  Ye, and Tie{-}Yan Liu. 2017.
\newblock \href
  {https://proceedings.neurips.cc/paper/2017/hash/6449f44a102fde848669bdd9eb6b76fa-Abstract.html}
  {Lightgbm: {A} highly efficient gradient boosting decision tree}.
\newblock In \emph{Advances in Neural Information Processing Systems 30: Annual
  Conference on Neural Information Processing Systems 2017, December 4-9, 2017,
  Long Beach, CA, {USA}}, pages 3146--3154.

\bibitem[{Li et~al.(2021)Li, Lin, Xiao, and Zhu}]{DBLP:conf/aaai/LiLXZ21}
Yanyang Li, Ye~Lin, Tong Xiao, and Jingbo Zhu. 2021.
\newblock \href {https://ojs.aaai.org/index.php/AAAI/article/view/17572} {An
  efficient transformer decoder with compressed sub-layers}.
\newblock In \emph{Thirty-Fifth {AAAI} Conference on Artificial Intelligence,
  {AAAI} 2021, Thirty-Third Conference on Innovative Applications of Artificial
  Intelligence, {IAAI} 2021, The Eleventh Symposium on Educational Advances in
  Artificial Intelligence, {EAAI} 2021, Virtual Event, February 2-9, 2021},
  pages 13315--13323. {AAAI} Press.

\bibitem[{Li et~al.(2020)Li, Hu, Zhang, Xu, Jiang, Xiao, Zhu, Liu, and
  Li}]{li-etal-2020-learning}
Yinqiao Li, Chi Hu, Yuhao Zhang, Nuo Xu, Yufan Jiang, Tong Xiao, Jingbo Zhu,
  Tongran Liu, and Changliang Li. 2020.
\newblock \href {https://doi.org/10.18653/v1/2020.acl-main.592} {Learning
  architectures from an extended search space for language modeling}.
\newblock In \emph{Proceedings of the 58th Annual Meeting of the Association
  for Computational Linguistics}, pages 6629--6639, Online. Association for
  Computational Linguistics.

\bibitem[{Lin et~al.(2021)Lin, Li, Wang, Li, Du, Xiao, and
  Zhu}]{lin-etal-2021-weight}
Ye~Lin, Yanyang Li, Ziyang Wang, Bei Li, Quan Du, Tong Xiao, and Jingbo Zhu.
  2021.
\newblock \href {https://doi.org/10.18653/v1/2021.acl-long.162} {Weight
  distillation: Transferring the knowledge in neural network parameters}.
\newblock In \emph{Proceedings of the 59th Annual Meeting of the Association
  for Computational Linguistics and the 11th International Joint Conference on
  Natural Language Processing (Volume 1: Long Papers)}, pages 2076--2088,
  Online. Association for Computational Linguistics.

\bibitem[{Liu et~al.(2018)Liu, Zoph, Neumann, Shlens, Hua, Li, Fei{-}Fei,
  Yuille, Huang, and Murphy}]{DBLP:conf/eccv/LiuZNSHLFYHM18}
Chenxi Liu, Barret Zoph, Maxim Neumann, Jonathon Shlens, Wei Hua, Li{-}Jia Li,
  Li~Fei{-}Fei, Alan~L. Yuille, Jonathan Huang, and Kevin Murphy. 2018.
\newblock \href {https://doi.org/10.1007/978-3-030-01246-5\_2} {Progressive
  neural architecture search}.
\newblock In \emph{Computer Vision - {ECCV} 2018 - 15th European Conference,
  Munich, Germany, September 8-14, 2018, Proceedings, Part {I}}, volume 11205
  of \emph{Lecture Notes in Computer Science}, pages 19--35. Springer.

\bibitem[{Liu et~al.(2019)Liu, Simonyan, and Yang}]{DBLP:conf/iclr/LiuSY19}
Hanxiao Liu, Karen Simonyan, and Yiming Yang. 2019.
\newblock \href {https://openreview.net/forum?id=S1eYHoC5FX} {{DARTS:}
  differentiable architecture search}.
\newblock In \emph{7th International Conference on Learning Representations,
  {ICLR} 2019, New Orleans, LA, USA, May 6-9, 2019}. OpenReview.net.

\bibitem[{{Long} et~al.(2019){Long}, {Zhang}, and {Zhang}}]{8901943}
D.~{Long}, S.~{Zhang}, and Y.~{Zhang}. 2019.
\newblock \href {https://doi.org/10.1109/CCHI.2019.8901943} {Performance
  prediction based on neural architecture features}.
\newblock In \emph{2019 2nd China Symposium on Cognitive Computing and Hybrid
  Intelligence (CCHI)}, pages 77--80.

\bibitem[{Lu et~al.(2019)Lu, Jiang, Xu, Shi, and
  Hu}]{DBLP:journals/corr/abs-1911-00105}
Qing Lu, Weiwen Jiang, Xiaowei Xu, Yiyu Shi, and Jingtong Hu. 2019.
\newblock \href {http://arxiv.org/abs/1911.00105} {On neural architecture
  search for resource-constrained hardware platforms}.
\newblock \emph{CoRR}, abs/1911.00105.

\bibitem[{Merity et~al.(2017)Merity, Xiong, Bradbury, and
  Socher}]{Merity2017PointerSM}
Stephen Merity, Caiming Xiong, James Bradbury, and Richard Socher. 2017.
\newblock \href {https://openreview.net/forum?id=Byj72udxe} {Pointer sentinel
  mixture models}.
\newblock In \emph{5th International Conference on Learning Representations,
  {ICLR} 2017, Toulon, France, April 24-26, 2017, Conference Track
  Proceedings}. OpenReview.net.

\bibitem[{Ning et~al.(2020)Ning, Zheng, Zhao, Wang, and
  Yang}]{DBLP:conf/eccv/NingZZWY20}
Xuefei Ning, Yin Zheng, Tianchen Zhao, Yu~Wang, and Huazhong Yang. 2020.
\newblock \href {https://doi.org/10.1007/978-3-030-58601-0\_12} {A generic
  graph-based neural architecture encoding scheme for predictor-based {NAS}}.
\newblock In \emph{Computer Vision - {ECCV} 2020 - 16th European Conference,
  Glasgow, UK, August 23-28, 2020, Proceedings, Part {XIII}}, volume 12358 of
  \emph{Lecture Notes in Computer Science}, pages 189--204. Springer.

\bibitem[{Pham et~al.(2018)Pham, Guan, Zoph, Le, and
  Dean}]{Pham2018EfficientNA}
Hieu Pham, Melody~Y. Guan, Barret Zoph, Quoc~V. Le, and Jeff Dean. 2018.
\newblock \href {http://proceedings.mlr.press/v80/pham18a.html} {Efficient
  neural architecture search via parameter sharing}.
\newblock In \emph{Proceedings of the 35th International Conference on Machine
  Learning, {ICML} 2018, Stockholmsm{\"{a}}ssan, Stockholm, Sweden, July 10-15,
  2018}, volume~80 of \emph{Proceedings of Machine Learning Research}, pages
  4092--4101. {PMLR}.

\bibitem[{Pham and Le(2021)}]{DBLP:conf/aaai/PhamL21}
Hieu Pham and Quoc~V. Le. 2021.
\newblock \href {https://ojs.aaai.org/index.php/AAAI/article/view/17127}
  {Autodropout: Learning dropout patterns to regularize deep networks}.
\newblock In \emph{Thirty-Fifth {AAAI} Conference on Artificial Intelligence,
  {AAAI} 2021, Thirty-Third Conference on Innovative Applications of Artificial
  Intelligence, {IAAI} 2021, The Eleventh Symposium on Educational Advances in
  Artificial Intelligence, {EAAI} 2021, Virtual Event, February 2-9, 2021},
  pages 9351--9359. {AAAI} Press.

\bibitem[{Press et~al.(2020)Press, Smith, and Levy}]{DBLP:conf/acl/PressSL20}
Ofir Press, Noah~A. Smith, and Omer Levy. 2020.
\newblock \href {https://doi.org/10.18653/v1/2020.acl-main.270} {Improving
  transformer models by reordering their sublayers}.
\newblock In \emph{Proceedings of the 58th Annual Meeting of the Association
  for Computational Linguistics}, pages 2996--3005, Online. Association for
  Computational Linguistics.

\bibitem[{Real et~al.(2019)Real, Aggarwal, Huang, and Le}]{2018regularized}
Esteban Real, Alok Aggarwal, Yanping Huang, and Quoc~V. Le. 2019.
\newblock \href {https://doi.org/10.1609/aaai.v33i01.33014780} {Regularized
  evolution for image classifier architecture search}.
\newblock In \emph{The Thirty-Third {AAAI} Conference on Artificial
  Intelligence, {AAAI} 2019, The Thirty-First Innovative Applications of
  Artificial Intelligence Conference, {IAAI} 2019, The Ninth {AAAI} Symposium
  on Educational Advances in Artificial Intelligence, {EAAI} 2019, Honolulu,
  Hawaii, USA, January 27 - February 1, 2019}, pages 4780--4789. {AAAI} Press.

\bibitem[{Shaw et~al.(2018)Shaw, Uszkoreit, and
  Vaswani}]{DBLP:conf/naacl/ShawUV18}
Peter Shaw, Jakob Uszkoreit, and Ashish Vaswani. 2018.
\newblock \href {https://doi.org/10.18653/v1/N18-2074} {Self-attention with
  relative position representations}.
\newblock In \emph{Proceedings of the 2018 Conference of the North {A}merican
  Chapter of the Association for Computational Linguistics: Human Language
  Technologies, Volume 2 (Short Papers)}, pages 464--468, New Orleans,
  Louisiana. Association for Computational Linguistics.

\bibitem[{So et~al.(2019)So, Le, and Liang}]{DBLP:conf/icml/SoLL19}
David~R. So, Quoc~V. Le, and Chen Liang. 2019.
\newblock \href {http://proceedings.mlr.press/v97/so19a.html} {The evolved
  transformer}.
\newblock In \emph{Proceedings of the 36th International Conference on Machine
  Learning, {ICML} 2019, 9-15 June 2019, Long Beach, California, {USA}},
  volume~97 of \emph{Proceedings of Machine Learning Research}, pages
  5877--5886. {PMLR}.

\bibitem[{Tan et~al.(2019)Tan, Chen, Pang, Vasudevan, Sandler, Howard, and
  Le}]{dblp:conf/cvpr/tancpvshl19}
Mingxing Tan, Bo~Chen, Ruoming Pang, Vijay Vasudevan, Mark Sandler, Andrew
  Howard, and Quoc~V. Le. 2019.
\newblock \href {https://doi.org/10.1109/CVPR.2019.00293} {Mnasnet:
  Platform-aware neural architecture search for mobile}.
\newblock In \emph{{IEEE} Conference on Computer Vision and Pattern
  Recognition, {CVPR} 2019, Long Beach, CA, USA, June 16-20, 2019}, pages
  2820--2828. Computer Vision Foundation / {IEEE}.

\bibitem[{Vaswani et~al.(2017)Vaswani, Shazeer, Parmar, Uszkoreit, Jones,
  Gomez, Kaiser, and Polosukhin}]{DBLP:conf/nips/VaswaniSPUJGKP17}
Ashish Vaswani, Noam Shazeer, Niki Parmar, Jakob Uszkoreit, Llion Jones,
  Aidan~N. Gomez, Lukasz Kaiser, and Illia Polosukhin. 2017.
\newblock \href
  {https://proceedings.neurips.cc/paper/2017/hash/3f5ee243547dee91fbd053c1c4a845aa-Abstract.html}
  {Attention is all you need}.
\newblock In \emph{Advances in Neural Information Processing Systems 30: Annual
  Conference on Neural Information Processing Systems 2017, December 4-9, 2017,
  Long Beach, CA, {USA}}, pages 5998--6008.

\bibitem[{Wang et~al.(2020)Wang, Wu, Liu, Cai, Zhu, Gan, and
  Han}]{wang2020hatht}
Hanrui Wang, Zhanghao Wu, Zhijian Liu, Han Cai, Ligeng Zhu, Chuang Gan, and
  Song Han. 2020.
\newblock \href {https://doi.org/10.18653/v1/2020.acl-main.686} {{HAT}:
  Hardware-aware transformers for efficient natural language processing}.
\newblock In \emph{Proceedings of the 58th Annual Meeting of the Association
  for Computational Linguistics}, pages 7675--7688, Online. Association for
  Computational Linguistics.

\bibitem[{Wang et~al.(2019)Wang, Li, Xiao, Zhu, Li, Wong, and
  Chao}]{wang-etal-2019-learning-deep}
Qiang Wang, Bei Li, Tong Xiao, Jingbo Zhu, Changliang Li, Derek~F. Wong, and
  Lidia~S. Chao. 2019.
\newblock \href {https://doi.org/10.18653/v1/P19-1176} {Learning deep
  transformer models for machine translation}.
\newblock In \emph{Proceedings of the 57th Annual Meeting of the Association
  for Computational Linguistics}, pages 1810--1822, Florence, Italy.
  Association for Computational Linguistics.

\bibitem[{Wauthier et~al.(2013)Wauthier, Jordan, and
  Jojic}]{DBLP:conf/icml/WauthierJJ13}
Fabian~L. Wauthier, Michael~I. Jordan, and Nebojsa Jojic. 2013.
\newblock \href {http://proceedings.mlr.press/v28/wauthier13.html} {Efficient
  ranking from pairwise comparisons}.
\newblock In \emph{Proceedings of the 30th International Conference on Machine
  Learning, {ICML} 2013, Atlanta, GA, USA, 16-21 June 2013}, volume~28 of
  \emph{{JMLR} Workshop and Conference Proceedings}, pages 109--117. JMLR.org.

\bibitem[{Wei et~al.(2020)Wei, Niu, Tang, and
  Liang}]{DBLP:journals/corr/abs-2003-12857}
Chen Wei, Chuang Niu, Yiping Tang, and Jimin Liang. 2020.
\newblock \href {http://arxiv.org/abs/2003.12857} {{NPENAS:} neural predictor
  guided evolution for neural architecture search}.
\newblock \emph{CoRR}, abs/2003.12857.

\bibitem[{Wen et~al.(2020)Wen, Liu, Chen, Li, Bender, and
  Kindermans}]{DBLP:conf/eccv/WenLCLBK20}
Wei Wen, Hanxiao Liu, Yiran Chen, Hai~Helen Li, Gabriel Bender, and
  Pieter{-}Jan Kindermans. 2020.
\newblock \href {https://doi.org/10.1007/978-3-030-58526-6\_39} {Neural
  predictor for neural architecture search}.
\newblock In \emph{Computer Vision - {ECCV} 2020 - 16th European Conference,
  Glasgow, UK, August 23-28, 2020, Proceedings, Part {XXIX}}, volume 12374 of
  \emph{Lecture Notes in Computer Science}, pages 660--676. Springer.

\bibitem[{Wu et~al.(2019{\natexlab{a}})Wu, Dai, Zhang, Wang, Sun, Wu, Tian,
  Vajda, Jia, and Keutzer}]{dblp:journals/corr/abs-1812-03443}
Bichen Wu, Xiaoliang Dai, Peizhao Zhang, Yanghan Wang, Fei Sun, Yiming Wu,
  Yuandong Tian, Peter Vajda, Yangqing Jia, and Kurt Keutzer.
  2019{\natexlab{a}}.
\newblock \href {https://doi.org/10.1109/CVPR.2019.01099} {Fbnet:
  Hardware-aware efficient convnet design via differentiable neural
  architecture search}.
\newblock In \emph{{IEEE} Conference on Computer Vision and Pattern
  Recognition, {CVPR} 2019, Long Beach, CA, USA, June 16-20, 2019}, pages
  10734--10742. Computer Vision Foundation / {IEEE}.

\bibitem[{Wu et~al.(2019{\natexlab{b}})Wu, Fan, Baevski, Dauphin, and
  Auli}]{DBLP:conf/iclr/WuFBDA19}
Felix Wu, Angela Fan, Alexei Baevski, Yann~N. Dauphin, and Michael Auli.
  2019{\natexlab{b}}.
\newblock \href {https://openreview.net/forum?id=SkVhlh09tX} {Pay less
  attention with lightweight and dynamic convolutions}.
\newblock In \emph{7th International Conference on Learning Representations,
  {ICLR} 2019, New Orleans, LA, USA, May 6-9, 2019}. OpenReview.net.

\bibitem[{Xiong et~al.(2020)Xiong, Yang, He, Zheng, Zheng, Xing, Zhang, Lan,
  Wang, and Liu}]{DBLP:conf/icml/XiongYHZZXZLWL20}
Ruibin Xiong, Yunchang Yang, Di~He, Kai Zheng, Shuxin Zheng, Chen Xing,
  Huishuai Zhang, Yanyan Lan, Liwei Wang, and Tie{-}Yan Liu. 2020.
\newblock \href {http://proceedings.mlr.press/v119/xiong20b.html} {On layer
  normalization in the transformer architecture}.
\newblock In \emph{Proceedings of the 37th International Conference on Machine
  Learning, {ICML} 2020, 13-18 July 2020, Virtual Event}, volume 119 of
  \emph{Proceedings of Machine Learning Research}, pages 10524--10533. {PMLR}.

\bibitem[{Zhao et~al.(2021)Zhao, Dong, Shen, Zhang, Wei, and
  Chen}]{DBLP:conf/acl/ZhaoDSZWC21}
Yuekai Zhao, Li~Dong, Yelong Shen, Zhihua Zhang, Furu Wei, and Weizhu Chen.
  2021.
\newblock \href {https://doi.org/10.18653/v1/2021.findings-acl.372}
  {Memory-efficient differentiable transformer architecture search}.
\newblock In \emph{Findings of the Association for Computational Linguistics:
  {ACL/IJCNLP} 2021, Online Event, August 1-6, 2021}, volume {ACL/IJCNLP} 2021
  of \emph{Findings of {ACL}}, pages 4254--4264. Association for Computational
  Linguistics.

\bibitem[{Zhong et~al.(2018)Zhong, Yan, Wu, Shao, and
  Liu}]{DBLP:conf/cvpr/ZhongYWSL18}
Zhao Zhong, Junjie Yan, Wei Wu, Jing Shao, and Cheng{-}Lin Liu. 2018.
\newblock \href {https://doi.org/10.1109/CVPR.2018.00257} {Practical block-wise
  neural network architecture generation}.
\newblock In \emph{2018 {IEEE} Conference on Computer Vision and Pattern
  Recognition, {CVPR} 2018, Salt Lake City, UT, USA, June 18-22, 2018}, pages
  2423--2432. {IEEE} Computer Society.

\bibitem[{Zoph et~al.(2018)Zoph, Vasudevan, Shlens, and
  Le}]{dblp:conf/cvpr/zophvsl18}
Barret Zoph, Vijay Vasudevan, Jonathon Shlens, and Quoc~V. Le. 2018.
\newblock \href {https://doi.org/10.1109/CVPR.2018.00907} {Learning
  transferable architectures for scalable image recognition}.
\newblock In \emph{2018 {IEEE} Conference on Computer Vision and Pattern
  Recognition, {CVPR} 2018, Salt Lake City, UT, USA, June 18-22, 2018}, pages
  8697--8710. {IEEE} Computer Society.

\end{thebibliography}
\bibliographystyle{acl_natbib}

\clearpage
\appendix

\section{Appendix}
\subsection{High-Accuracy Architecture Search Space}
\label{appendix-space}

Other design choices are adopted from HAT's search space \cite{wang2020hatht} with slight modifications. Inspired by \citet{DBLP:conf/naacl/ShawUV18}, we search for the maximum relative position (\textit{RPR Len}) in the self-attention modules in each layer. As suggested by \citet{wang-etal-2019-learning-deep} and \citet{DBLP:conf/icml/XiongYHZZXZLWL20}, proper locations of layer normalization lead to better performance. Therefore, we let NAS decide whether to put the layer normalization inside (Pre-LN) or between (Post-LN) the residual blocks.

\begin{table}[ht]
\centering
\scalebox{1}{
\begin{tabular}{c|c}
\toprule
Features & Search Space \\ 
\midrule
\textit{Enc Layer Num}  &   [6] \\
\textit{Enc Emb Dim}    &   [512, 640, 768] \\
\textit{Enc FFN Dim}    &   [768, 1024, 1536, 2048] \\
\textit{Enc Head Num}   &   [2, 4, 8] \\
\textit{Enc RPR Len}    &   [8, 12, 16] \\
\textit{Enc Norm Type}  &   [Pre-LN, Post-LN] \\
\textit{Dec Layer Num}  &   [1, 2, 3, 4, 5, 6] \\
\textit{Dec Emb Dim}    &   [512, 640, 768] \\
\textit{Dec FFN Dim}    &   [768, 1024, 1536, 2048] \\
\textit{Dec Head Num}   &   [2, 4, 8] \\
\textit{Dec RPR Len}    &   [8, 12, 16] \\
\textit{Dec Norm Type}  &   [Pre-LN, Post-LN] \\
\textit{Enc-Dec Attn}   &   [1, 2, 3] \\
\midrule
\end{tabular}
}
\caption{The search space for high-accuracy search on the IWSLT'14 De-En translation task.}
\label{tab:iwslt-search-space}
\end{table}
\begin{table}[ht]
\centering
\scalebox{1}{
\begin{tabular}{c|c}
\toprule
Features & Search Space \\ 
\midrule
\textit{Enc Layer Num}  &   [6] \\
\textit{Enc Emb Dim}    &   [640, 768, 1024] \\
\textit{Enc FFN Dim}    &   [2048, 3072, 4096, 5120] \\
\textit{Enc Head Num}   &   [4, 8, 16] \\
\textit{Enc RPR Len}    &   [8, 12, 16] \\
\textit{Enc Norm Type}  &   [Pre-LN, Post-LN] \\
\textit{Dec Layer Num}  &   [1, 2, 3, 4, 5, 6] \\
\textit{Dec Emb Dim}    &   [640, 768, 1024] \\
\textit{Dec FFN Dim}    &   [2048, 3072, 4096, 5120] \\
\textit{Dec Head Num}   &   [4, 8, 16] \\
\textit{Dec RPR Len}    &   [8, 12, 16] \\
\textit{Dec Norm Type}  &   [Pre-LN, Post-LN] \\
\textit{Enc-Dec Attn}   &   [1, 2, 3] \\
\midrule
\end{tabular}
}
\caption{The search space for high-accuracy search on the WMT'14 En-De translation task.}
\label{tab:wmt-search-space}
\end{table}

\subsection{Visualization of Good Architectures}
\label{appendix-arch}

\begin{figure}[t!]
    \centering
    \includegraphics[scale=0.5]{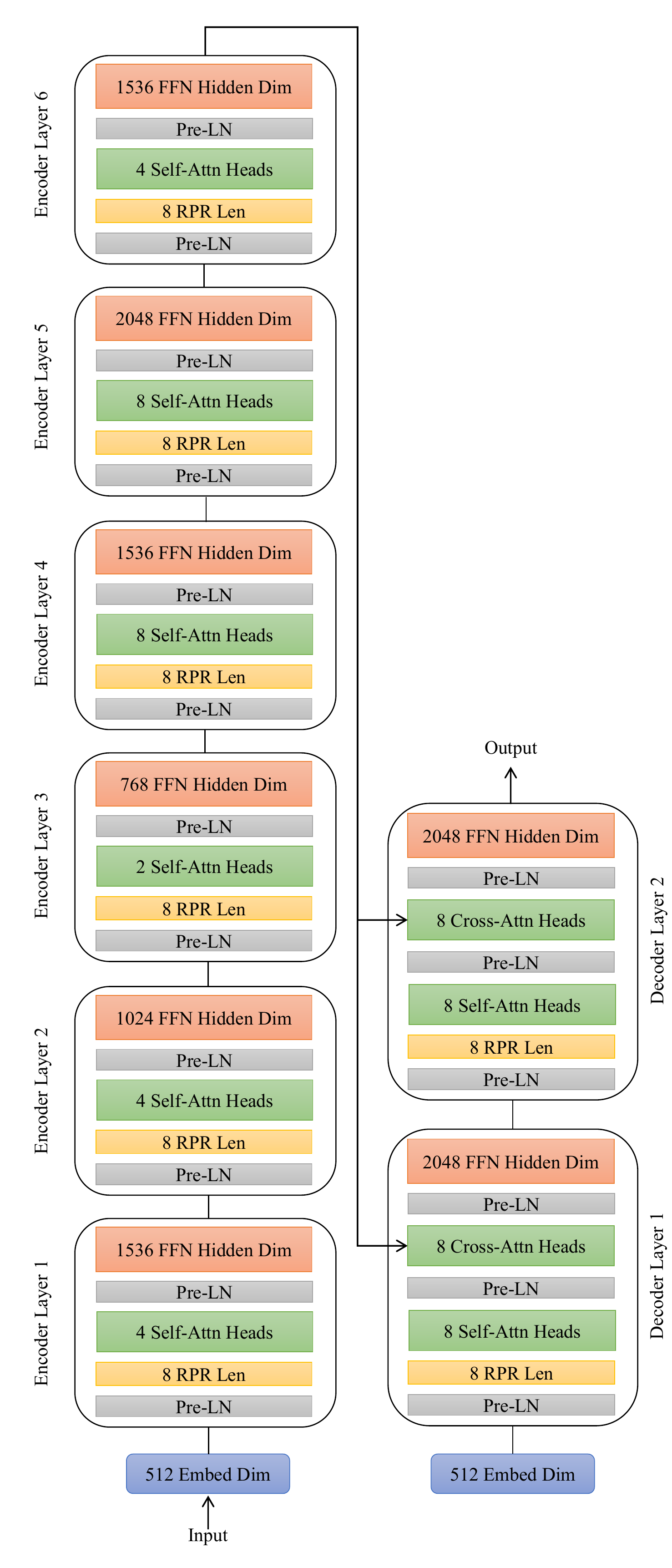}
    \caption{Visualization of a discovered architecture on the IWSLT'14 De-En translation task.}
    \label{fig:iwslt-arch}
\end{figure}

Figure \ref{fig:iwslt-arch} illustrates one of the discovered Transformer architecture. The presented architecture achieves 36.2 BLEU on the IWSLT'14 De-En translation task and has a latency of 77ms on the GTX 1080Ti GPU, outperforming the vanilla Transformer by +1.8 BLEU and 2.6 times speed.

\end{document}